\documentclass[pdflatex,sn-mathphys,iicol]{sn-jnl}

\jyear{2023}%

\theoremstyle{thmstyleone}%
\theoremstyle{thmstyletwo}%

\theoremstyle{thmstylethree}%

\usepackage{url}

\usepackage{graphicx}
\usepackage{amsmath,amssymb} 
\usepackage{color}
\usepackage{booktabs}

\usepackage{multirow}
\usepackage[normalem]{ulem}
\useunder{\uline}{\ul}{}

\usepackage{threeparttable}

\usepackage{hhline}
\usepackage{siunitx}
\usepackage{booktabs}
\usepackage{subcaption}

\usepackage{colortbl}
\usepackage{arydshln}  
\PassOptionsToPackage{hyphens}{url}\usepackage{hyperref}

\begin{document}

\newcommand{\etal}{\textit{et al.}}


\title[Exploiting Autoencoder's Weakness to Generate Pseudo Anomalies]{Exploiting Autoencoder's Weakness to Generate Pseudo Anomalies}


\author[1,2,3]{\fnm{Marcella} \sur{Astrid}}\email{marcella.astrid@uni.lu}

\author[4]{\fnm{Muhammad Zaigham} \sur{Zaheer}}\email{zaigham.zaheer@mbzuai.ac.ae}

\author[3]{\fnm{Djamila} \sur{Aouada}}\email{djamila.aouada@uni.lu}

\author*[1,2]{\fnm{Seung-Ik} \sur{Lee}}\email{the\_silee@etri.re.kr}

\affil[1]{\orgdiv{Department of Artificial Intelligence}, \orgname{University of Science and Technology}, \orgaddress{\city{Daejeon}, \postcode{34113}, \country{South Korea}}}

\affil[2]{\orgdiv{Field Robotics Research Section}, \orgname{Electronics and Telecommunications Research Institute}, \orgaddress{\city{Daejeon}, \postcode{34129}, \country{South Korea}}}

\affil[3]{\orgdiv{Interdisciplinary Centre for Security, Reliability and Trust}, \orgname{University of Luxembourg}, \orgaddress{\city{Luxembourg}, \postcode{1855}, \country{Luxembourg}}}

\affil[4]{\orgdiv{Department of Computer Vision}, \orgname{Mohamed Bin Zayed University of Artificial Intelligence}, \orgaddress{\city{Abu Dhabi}, \country{United Arab Emirates}}}


\abstract{
  Due to the rare occurrence of anomalous events, a typical approach to anomaly detection is to train an autoencoder (AE) with normal data only so that it learns the patterns or representations of the normal training data. 
  At test time, the trained AE is expected to well reconstruct normal but to poorly reconstruct anomalous data. However, contrary to the expectation, anomalous data is often well reconstructed as well. 
  In order to further separate the reconstruction quality between normal and anomalous data, we propose creating pseudo anomalies from learned adaptive noise by exploiting the aforementioned weakness of AE, i.e., reconstructing anomalies too well. 
  The generated noise is added to the normal data to create pseudo anomalies.
  Extensive experiments on Ped2, Avenue, ShanghaiTech, CIFAR-10, and KDDCUP datasets demonstrate the effectiveness and generic applicability of our approach in improving the discriminative capability of AEs for anomaly detection.}




\maketitle
\section{Introduction}\label{sec1}

\pagestyle{plain}%

Anomaly detection has recently garnered significant attention from numerous researchers due to its pivotal role in various applications, such as automatic surveillance systems \cite{zaheer2020old,gong2019memorizing,li2022anomaly,astrid2021learning,sultani2018real}, medical data analysis \cite{wahid2021nanod,abhaya2022rdpod,vafaeiflexible}, network intrusion prevention \cite{zavrak2023flow,shukla2021detection,saeed2022real}.
By definition, anomalous events are rare and can be cumbersome to collect. Therefore, anomaly detection is commonly approached as a one-class classification (OCC) problem, in which only normal data is utilized to train a detection model. 

\begin{figure*}[t]
\begin{center}
\includegraphics[width=\linewidth]{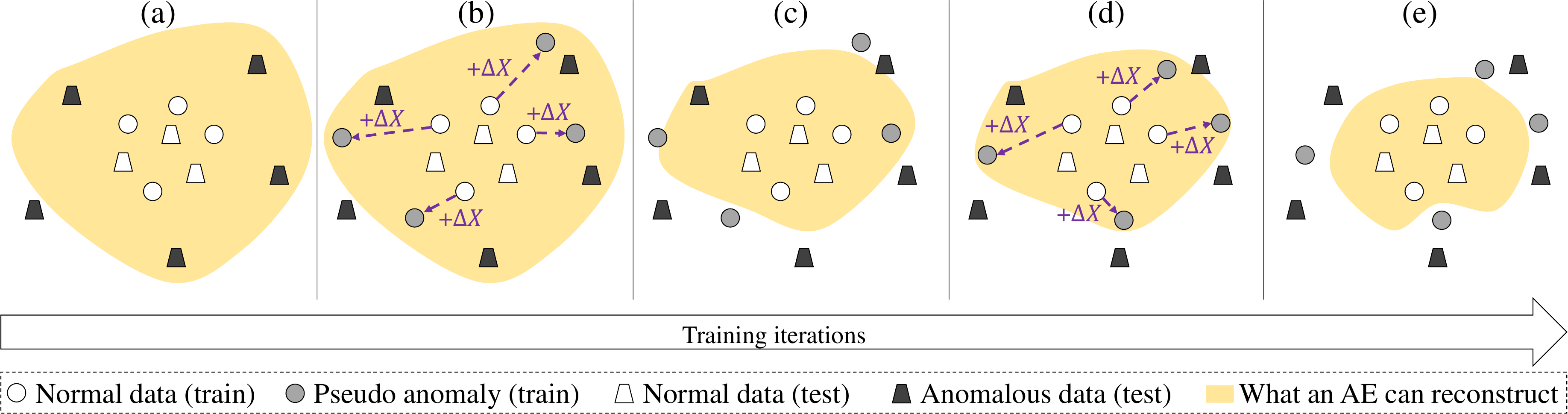}
\end{center}
   \caption{Illustration on how our method limits the reconstruction capability of an AE across training iterations: (a) AE can reconstruct both normal data and anomalous data, (b) The noise generator generates a noise $\Delta X$ to produce pseudo anomalies within the reconstruction boundary of AE, (c) AE learns to poorly reconstruct pseudo anomalies, (d) Pseudo anomalies are generated to adapt to the new reconstruction boundary, and (e) AE learns to poorly reconstruct the new pseudo anomalies.}
\label{fig:pseudoanomalylearnednoise_illustration}
\end{figure*}

Typically, an autoencoder (AE) model is employed to address the OCC problem 
\cite{luo2017revisit,luo2017remembering,gong2019memorizing,park2020learning,astrid2021learning,astrid2021synthetic,liu2023csiamese}. 
By learning to reconstruct the normal training data, an AE encodes representations of normalcy in its latent space.
During testing, the model is expected to reconstruct normal data well but exhibit poor reconstruction in anomalous data. However, as observed by  \cite{munawar2017limiting,gong2019memorizing,zaheer2020old,astrid2021learning,astrid2021synthetic}, AEs suffer from reconstructing anomalous data too well, leading to reduced discrimination between normal and anomalous data, as illustrated in Figure \ref{fig:pseudoanomalylearnednoise_illustration}(a).

To address this phenomenon, \cite{gong2019memorizing,park2020learning} employ memory-based networks to memorize normal patterns in the latent space.
This compels an AE to utilize the memorized latent codes for input reconstruction.
These approaches have proven successful in ensuring poor reconstruction for anomalous data. 
However, a potential issue arises in that it may also restrict the reconstruction of normal data, depending on the memory size, as depicted in Figure 6 of \cite{gong2019memorizing}.

\begin{figure*}
\begin{center}
\includegraphics[width=\linewidth]{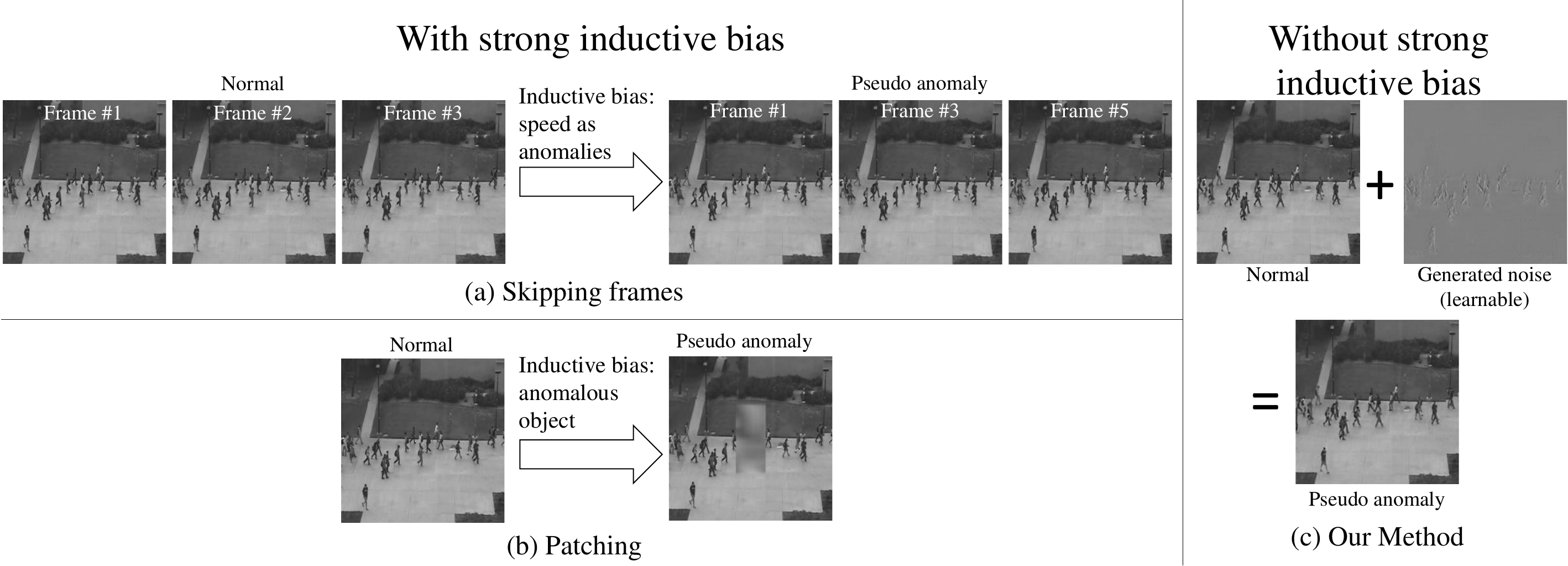}
\end{center}
   \caption{
   Comparison of pseudo anomaly generation using (a) skipping frames \cite{astrid2021learning,astrid2021synthetic}, (b) patching \cite{astrid2021learning}, and (c) our method. Our method is learnable and does not impose any strong inductive bias.}
\label{fig:introduction}

\end{figure*}

To enable an AE to learn more appropriate reconstruction boundary, recently, \cite{astrid2021learning,astrid2021synthetic} proposed the utilization of pseudo anomalies to assist the training of an AE. Pseudo anomalies are fake anomalies generated from normal data in the training set to emulate anomalous data. The AE is subsequently trained using both normal and pseudo anomalous data. 
While the model is trained with normal data to minimize the reconstruction loss, in the case of pseudo anomalous data, the AE is trained to inadequately reconstruct the input.
Despite these methods outperform memory-based networks, one notable drawback is that pseudo anomalies are synthesized on the assumptions, such as the speed of anomalous movements \cite{astrid2021learning,astrid2021synthetic} (Figure \ref{fig:introduction}(a)) or anomalous objects \cite{astrid2021learning} (Figure \ref{fig:introduction}(b)), significantly limiting their applicability.

To avoid making such assumptions and, more importantly, to achieve greater generality across diverse areas,
we propose constructing pseudo anomalies by incorporating learned adaptive noise to the normal data (Figure \ref{fig:introduction}(c)).
To implement this, we simultaneously train another network that learns to generate noise based on a normal input. Pseudo anomaly data are subsequently created by adding the generated noise to the input (Figure \ref{fig:pseudoanomalylearnednoise_illustration}(b) \& Figure \ref{fig:pseudoanomalylearnednoise_illustration}(d)). An AE is then trained to poorly reconstruct the generated pseudo anomalies (Figure \ref{fig:pseudoanomalylearnednoise_illustration}(c) \& Figure \ref{fig:pseudoanomalylearnednoise_illustration}(e)). This approach ensures that the reconstruction boundary of the AE may evolve towards a significantly improved one as the training progresses, as illustrated in Figure \ref{fig:pseudoanomalylearnednoise_illustration}.

There has been a controversy among machine learning researchers regarding whether it limits the idea of artificial general intelligence \cite{voss2007essentials} to impose a strong inductive bias or in other words, to make use of strong assumptions such as in \cite{astrid2021learning,astrid2021synthetic} for an improved performance. 
Undoubtedly, inductive bias can be beneficial for highly specific practical anomaly detection solutions. 
However, methods relying on such bias can be vulnerable in cases where the underlying assumptions do not hold. Additionally, their applicability is extremely limited to specific applications, e.g., video. 
On the other hand, our work refrains from employing any strong inductive bias, making it generic and applicable across diverse domains, from visual (videos and images) to network intrusion data.

In summary, the contributions of our work are as follows: 
1) Our work is among the first few to explore the possibility of generating pseudo anomalies in anomaly detection;
2) We leverage the well-known weakness of AE, i.e., their tendency to reconstruct anomalies too well, to our advantage. We achieve this by learning to generate pseudo anomalies that can hinder the successful reconstruction of anomalies;
3) Our noise-based pseudo anomaly generation assists in training without relying on any strong inductive bias;
4) We extensively evaluate our method, demonstrating its broad applicability across highly complex video, image, and network intrusion datasets: Ped2 \cite{li2013anomaly}, Avenue \cite{lu2013abnormal}, ShanghaiTech \cite{luo2017revisit}, CIFAR-10 \cite{krizhevsky2009learning}, and KDDCUP \cite{Dua:2019}.  

\section{Related work}
\label{sec:relatedworks}

\subsection{Limiting the reconstruction of AE} 
\label{subsec:limitingreconAE}

To address the issue of AEs reconstructing anomalies too well, 
\cite{gong2019memorizing,park2020learning} introduce memory mechanisms into the latent space to constrain the reconstruction capability of an AE. However, this approach may inadvertently restrict normal data reconstructions as well. Another strategy proposed by \cite{astrid2021learning,astrid2021synthetic,astrid2023pseudobound} involves utilizing data-heuristic pseudo anomalies. 
When pseudo anomalies are input, \cite{astrid2021synthetic,astrid2023pseudobound} aim to maximize the reconstruction loss, while \cite{astrid2021learning} minimizes the reconstruction loss with respect to the normal data used for generating the pseudo anomalies.
Essentially, our training configuration is similar to that of \cite{astrid2021learning}, as we also minimize the reconstruction loss with respect to the normal data. 
However, we distinguish our method by not explicitly imposing any inductive bias during pseudo anomaly generation, thereby extending the applicability of our approach to a broader range of applications.

\subsection{Pseudo anomalies} 

Several previous studies take advantage of inductive bias to generate pseudo anomalies. \cite{astrid2021learning, astrid2021synthetic} propose a skipping frame technique that considers the relatively fast movements of objects as anomalies. Similarly, \cite{astrid2021learning, zhong2022cascade} suggest putting a patch from another dataset into an image to mimic the existence of anomalous objects. In addition, to create pseudo anomalies, PseudoBound \cite{astrid2023pseudobound} explores several hand-crafted augmentation techniques, such as skipping frames, adding patches, and incorporating noise into the normal data. In contrast, our work does not impose any such inductive bias or assumption (e.g., fast movements, existence of anomalous objects) nor use hand-crafted augmentation techniques. Instead, we allow the model to learn to generate pseudo anomalies based on the reconstruction boundary of AEs.

\subsection{Adversarial training}
\label{subsec:adversarialtraining}
From an architectural point-of-view, our method of learning to generate pseudo anomalies may also be viewed as related to adversarial training \cite{goodfellow2014generative}. 
In this context, OGNet \cite{zaheer2020old,zaheer2022stabilizing} and G2D \cite{pourreza2021g2d} also employ adversarial training, where an AE serves as the generator and a binary classifier functions as the discriminator. In the second phase of training, pseudo anomalies obtained from an undertrained generator in the first phase are used to train the binary classifier. 
In contrast, we utilize the AE itself as discriminator, eliminating the need for a separate discriminator. 
Furthermore, unlike existing methods, ours learns to generate pseudo anomalies, thereby freeing us from the burden of designing any hand-crafted schemes for pseudo anomalies.

\subsection{Non-reconstruction methods} 
Different from our method, non-reconstruction methods do not utilize reconstruction as the training objective and/or as the sole decision factor of the anomaly score.
Several approaches fall into this category, such as predicting future frames \cite{liu2018future,park2020learning}, utilizing object detection under the assumption that anomalous events are always related to objects \cite{ionescu2019object,georgescu2021background}, adding optical flow components \cite{ji2020tam,lee2019bman}, and using a binary classifier to predict anomaly scores \cite{zaheer2020old,pourreza2021g2d}. Since some of these methods utilize AE \cite{georgescu2021background,park2020learning,liu2018future}, our work can also potentially be incorporated into these methods.

\subsection{Non-OCC methods} 
To enhance the discrimination capability of an AE, several researchers incorporate real anomalies during training \cite{munawar2017limiting,yamanaka2019autoencoding}. Recently, video-level weakly supervised \cite{sultani2018real,zaheer2020claws,zaheer2023clustering,karim2024real,majhi2024oe} or fully unsupervised \cite{zaheer2022generative} training configurations have also been introduced. 
However, due to the limited variety of anomalous data, these approaches are prone to overfitting and are restricted to specific anomalous cases appearing in the training data.
On the other hand, our method utilizes only normal training data, offering greater potential to build a highly generic model capable of working with diverse anomalous data that may deviate from the normal patterns in the training data.

\subsection{With vs. without inductive bias} 
In existing literature, several anomaly detection approaches impose inductive bias.
Object-centric methods \cite{ionescu2019detecting,georgescu2021background} assume that anomalous events are related to objects. However, such methods encounter difficulties in detecting non-object-based anomalous events, such as unattended fires or blasts. Several other pseudo-anomaly-based methods also make strong assumptions, such as fast movements \cite{astrid2021learning,astrid2021synthetic} or out-of-distribution objects as anomalies \cite{astrid2021learning,zhong2022cascade}. Consequently, their applicability may be severely limited, for example, in video data.
On the other hand, there have been anomaly detection methods that do not impose inductive bias, such as memory-based networks \cite{gong2019memorizing,park2020learning}, whose limitations are discussed in Section \ref{subsec:limitingreconAE}.

\subsection{Denoising AE} 
Our work is also related to denoising AE \cite{salehi2021arae,jewell2022one,vincent2008extracting}. However, the usage and purpose of the noise are where differ. A typical denoising AE is trained with random noise added to the input to capture robust features and to prevent the network from merely duplicating the input at the output. 
In contrast, we learn to generate noise to create pseudo anomalies. 
 

\subsection{Data augmentation}
Creating pseudo anomalies can be seen as data augmentation since we incorporate the generated pseudo anomalies during training, and they contribute to the loss. However, instead of adding data from the same classes as in the typical data augmentation techniques \cite{bengio2011deep,krizhevsky2012imagenet}, creating pseudo anomalies introduces a new class, i.e., the anomaly class, into the normal-only training set. Several data augmentation techniques also employ adversarial training to generate augmented examples that are adversarial for the model \cite{zhang2020adversarial,tang2020onlineaugment}. In such approaches, the goal is to discover augmented data that can fool the classification network into predicting the wrong category. 
Our method also utilizes adversarial training to generate pseudo anomalies. However, different from adversarial-based augmentation techniques, our method generates augmented data that the AE can reconstruct well.

\subsection{Generation of out-of-distribution data} 

Our method of generating pseudo anomalies is also related to the generation of out-of-distribution (OOD) data. In semi-supervised learning \cite{dai2017good,dong2019margingan}, OOD data are utilized to assist in training a classifier with a small amount of labeled data and a large amount of unlabeled data. Bad GAN \cite{dai2017good} obtains OOD data by employing PixelCNN++ \cite{salimans2017pixelcnn++}, an external separate network that predicts data density. Margin GAN \cite{dong2019margingan} proposes generating real-looking fake data that fools both the discriminator and the classifier. In object detection, VOS \cite{du2022towards} models object features in each class using a Gaussian distribution and then generate outliers from the Gaussian, allowing the classifier to learn to recognize objects outside the provided object classes.
In deepfake detection, FaceXRay \cite{fxray} and SBI \cite{sbis} assume the existence of mask boundary artifacts in the deepfakes to create OOD data as pseudo-fake.
In contrast, our method does not assume any particular family of density functions or the appearance of anomalous data to generate OOD samples. Instead, our model learns to generate OOD data near the reconstruction boundary without making assumptions about the nature of anomalous data.

\begin{figure*}[t]
\begin{center}
\includegraphics[width=\linewidth]{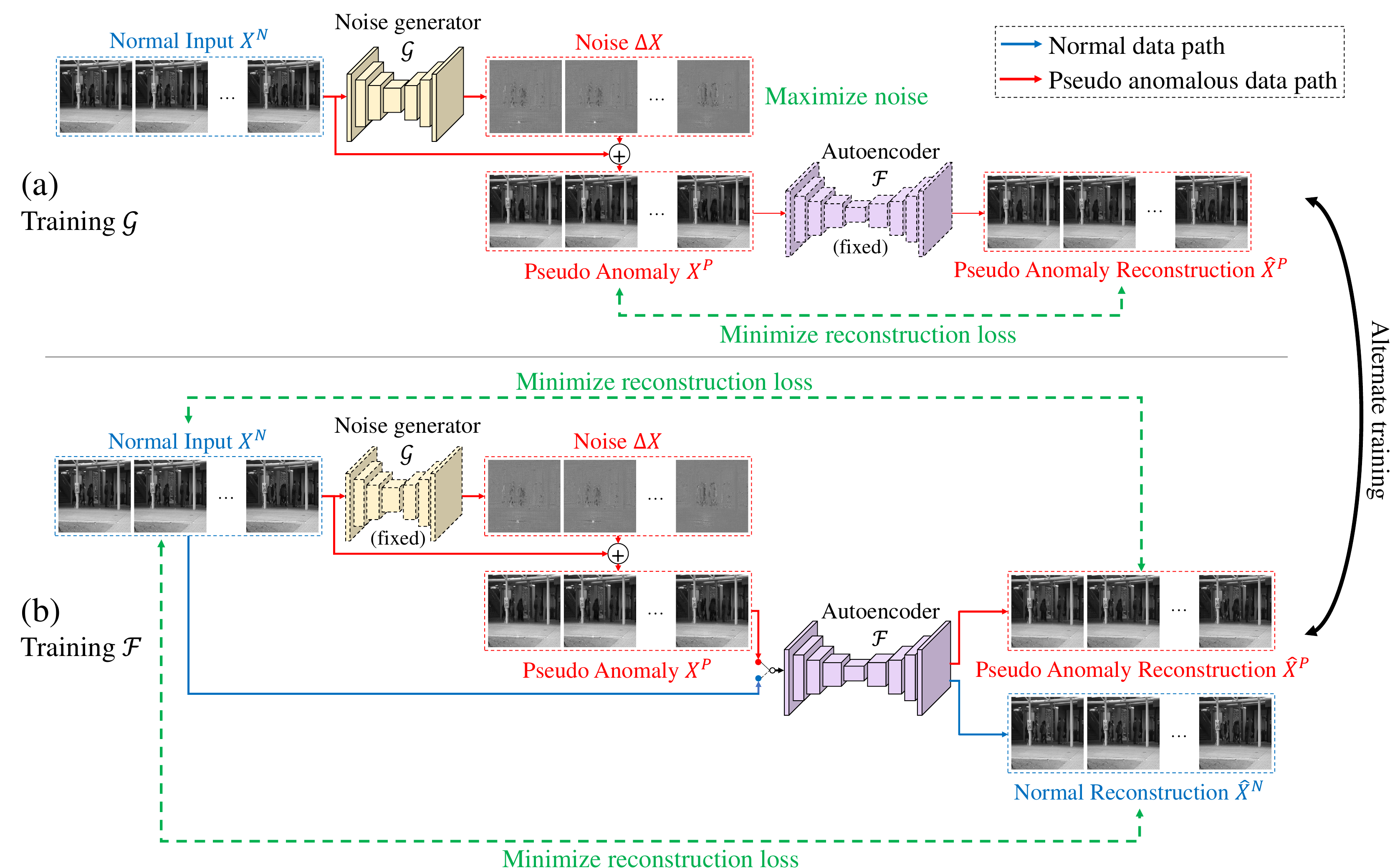}
\end{center}
   \caption{Our method consists of a main autoencoder $\mathcal{F}$ and a noise generator $\mathcal{G}$ that are  trained alternately: (a) A pseudo anomaly instance is constructed by adding noise to the normal data, where the noise is generated by $\mathcal{G}$. 
   $\mathcal{G}$ learns to generate as much noise as $\mathcal{F}$ is able to reconstruct the pseudo anomalies. In other words, 
   $\mathcal{G}$ is trained to generate anomalies (maximizing noise) that are within the reconstruction boundary of $\mathcal{F}$ (minimizing reconstruction loss). (b) $\mathcal{F}$ is trained to not reconstruct anomalies when the inputs are generated pseudo anomalies and trained to reconstruct normal data when the inputs are normal. During test time, only $\mathcal{F}$ is used. The contrast of $\Delta X$ has been adjusted for visualization clarity.}
\label{fig:overall}
\end{figure*}

\section{Methodology}
\label{sec:methodology}

In this section, we discuss our proposed approach of learning to generate pseudo anomalies. The overall configuration is illustrated in Figure \ref{fig:overall}, where an autoencoder $\mathcal{F}$ (Figure \ref{fig:overall}(b)) and a noise generator $\mathcal{G}$ (Figure \ref{fig:overall}(a)) are trained alternately.

\subsection{Learning not to reconstruct anomalies ($\mathcal{F}$)}
\label{subsec:trainF}

In the OCC setting, an autoencoder (AE) is typically employed as a reconstruction model \cite{hasan2016learning,zhao2017spatio,luo2017revisit,luo2017remembering,gong2019memorizing,park2020learning}. 
To enhance the discrimination between normal and anomalous reconstruction, we train the AE $\mathcal{F}$ using pseudo anomaly input $X^P$ with a probability $p$ and normal input $X^N$ with a probability $1-p$. The AE then outputs the reconstruction of normal data and pseudo anomalous data as:
\begin{equation}
    \hat{X}^N = \mathcal{F}(X^N) \text{,} 
\label{eq:ae}
\end{equation}
\begin{equation}
    \hat{X}^P = \mathcal{F}(X^P) \text{,}
\label{eq:aepseudo}
\end{equation}
respectively.

A pseudo anomaly $X^P$ is supposedly anomalous data generated from the normal training data. It is considered anomalous as it does not conform to the normalcy defined in the training data. 
Additionally, it is pseudo as it is not a real anomaly.
In this work, we propose to generate noise $\Delta X$ from a normal input $X^N$ to construct a pseudo anomaly $X^P$ by adding $\Delta X$ to $X^N$:
\begin{equation}
    X^P = X^N + \Delta X \text{.}
\label{eq:pseudoanomaly}
\end{equation}
To generate $\Delta X$, an additional autoencoder $\mathcal{G}$ is employed as:
\begin{equation}
    \Delta X = \mathcal{G}(X^N) \text{.}
\label{eq:pseudolearnable}
\end{equation}
The intuition behind generating $X^P$ from $X^N$ is illustrated in Figure \ref{fig:pseudoanomalylearnednoise_illustration}(b) \& (d).

In order to train $\mathcal{F}$ to well reconstruct normal input while poorly reconstructing anomalies, we aim to train $\mathcal{F}$ to reconstruct any given input (whether $X^N$ or $X^P$) as its normal data $X^N$. Therefore, during normal input, the training involves minimizing the reconstruction loss using mean squared error between $X^N$ and $\hat{X}^N$:
\begin{equation}
    \min_{\mathcal{F}}  \frac{1}{d} \left \| \hat{X}^N - X^N  \right \|_{F}^{2}  \text{,}
\label{eq:reconloss}
\end{equation}
where $\left \| .  \right \|_{F}$ denotes Frobenius norm and $d$ is the number of elements in $X^N$. Even when the input is a pseudo anomaly, the reconstruction loss is calculated with the corresponding normal data $X^N$ as the target:
\begin{equation}
    \min_{\mathcal{F}}  \frac{1}{d} \left \| \hat{X}^P - X^N  \right \|_{F}^{2}  \text{.}
\label{eq:pseudoreconloss}
\end{equation}
This way, $\mathcal{F}$ is learned not to reconstruct pseudo anomalies, as illustrated in Figure \ref{fig:pseudoanomalylearnednoise_illustration}(c) \& (e). The training mechanism of $\mathcal{F}$ can be seen in Figure \ref{fig:overall}(b).

\subsection{Learning to generate pseudo anomalies ($\mathcal{G}$)}
\label{subsec:trainG}

$\mathcal{G}$ is trained by exploiting what we may term as the weakness of $\mathcal{F}$ in anomaly detection, i.e., reconstructing too well on anomalous data. 
Therefore, we propose a loss consisting 
of two parts, including reconstruction and noise amplitude:
\begin{equation}
    \min_{\mathcal{G}}   \frac{1}{d} \left( \left \| \hat{X}^P - X^P  \right \|_{F}^{2} - \lambda \left \| \Delta X  \right \|_{F}^{2} \right) \text{,}
\label{eq:learningnoise}
\end{equation}
where $\lambda$ is a balancing hyperparameter.
The first part of the training objective is to reduce the reconstruction loss of $\mathcal{F}$ for the pseudo anomalous input. Notice that, different from Eq. \eqref{eq:pseudoreconloss}, the target for the reconstruction loss in Eq. \eqref{eq:learningnoise} is $X^P$. In this way, $\mathcal{G}$ is trained to create noise in such a way that the resultant pseudo anomaly is within the reconstruction boundary of $\mathcal{F}$.
On the other hand, the second part of the loss encourages the norm of the noise $\Delta X$ to be high which makes
the resultant pseudo anomaly to be far from the normal. 
In summary, the overall loss encourages $\mathcal{G}$ to generate highly noisy pseudo anomalies that can be reconstructed by $\mathcal{F}$ (Figure \ref{fig:pseudoanomalylearnednoise_illustration}(b) \& (d)).
The training of $\mathcal{G}$ is illustrated in Figure \ref{fig:overall}(a).
$\mathcal{G}$ is alternately trained with $\mathcal{F}$ and adapts to the reconstruction boundary of $\mathcal{F}$, as illustrated in Figure \ref{fig:pseudoanomalylearnednoise_illustration}.
It is pertinent to note that learning to generate noise for pseudo anomalies does not impose any inductive bias, which allows our model to have generic applicability. 

\subsection{Test time}
\label{subsec:testtime}

At inference, we compute the anomaly score using the reconstruction error of $\mathcal{F}$, where a higher reconstruction error corresponds to a higher anomaly score. Without using $\mathcal{G}$, we directly input the test data into $\mathcal{F}$ and calculate the reconstruction error by comparing the output with the input. In this way, our final model does not incur any additional computational cost compared to the baseline (i.e., AE trained only with normal data).

\begin{table*}[]
\caption{Default hyperparameter values used in this method for each setup and dataset.}
\small
\centering
\resizebox{\linewidth}{!}{
\begin{tabular}{|ccc|ccc|ccc|ccc|}
\hline
\multicolumn{3}{|c|}{Probability $p$} & \multicolumn{3}{c|}{Weighting factor $\lambda$} & \multicolumn{3}{c|}{Batch size} & \multicolumn{3}{c|}{$\mathcal{F}$ learning rate} \\ \hline
Video     & Image     & Network     & Video        & Image        & Network        & Video    & Image    & Network   & Video    & Image    & Network  \\ \hline
0.5       & 0.1       & 0.5         & 0.1          & 10           & 1              & 4        & 256      & 1024   & $10^{-4}$    & $10^{-3}$    & $10^{-4}$     \\ \hline
\end{tabular}
}
\label{tab:default_hyperparameters}
\end{table*}

\section{Experiments}
\label{sec:experiments}
In this section, we first define our datasets (Section \ref{subsec:datasets}) and experiment setup (Section \ref{subsec:experiment_setup}) used in the experiments. To evaluate the effectiveness of our approach in generating pseudo anomalies with learnable noise, we extensively compare the performances and properties between the baseline and our method in Section \ref{subsec:ablation}. We then compare the performance and generic applicability of our method with the state-of-the-art (SOTA) on various applications in Section \ref{subsec:comparisons_sota}. As an in-depth analysis, we explore the effects of the hyperparameters introduced in our method and discuss their selection in Section \ref{subsec:hyperparameter_evaluation}, and the design choices and training progression are discussed in Section \ref{subsec:additional_discussions}.

\subsection{Datasets}
\label{subsec:datasets}

We evaluate our method on several highly complex datasets including three surveillance video datasets (i.e., Ped2 \cite{li2013anomaly}, Avenue \cite{lu2013abnormal}, and ShanghaiTech \cite{luo2017revisit}) as well as an image dataset CIFAR-10 \cite{krizhevsky2009learning}. To further show the wide applicability of our proposed method, we also assess its performance on non-visual datasets such as the network security dataset, KDDCUP99 \cite{Dua:2019}. 
Details of each dataset used in our experiments are as follows:

\subsubsection{Ped2} It comprises 16 training and 12 test videos. The normal scenes consist of pedestrians only, whereas anomalous scenes include bikes, carts, or skateboards along with pedestrians.

\subsubsection{Avenue} It consists of 16 training and 21 test videos.
Examples of anomalies are abnormal objects, such as bikes, and abnormal actions of humans, such as unusual walking directions, running, or throwing objects.

\subsubsection{ShanghaiTech} This is by far the largest one-class anomaly detection dataset consisting of 330 training and 107 test videos.
The dataset is recorded at 13 different locations of complex lighting conditions and camera angles. Within the test videos, there are 130 anomalous events including running, riding bicycle, and fighting.

\subsubsection{CIFAR-10} It is originally a 10 classes image classification dataset. For anomaly detection experiments, we adopt the setup from \cite{abati2019latent} and set one class as normal while the others as anomaly. 
To train and evaluate our model, we adhere to the original split of the CIFAR-10 dataset into training and test sets. However, during training, we exclude the selected anomaly classes.
Similar to \cite{abati2019latent}, we also separate 10\% of the original training split for validation. 
We choose the best validation model as our final model for the test. 

\subsubsection{KDDCUP}
Following the setup employed in other methods \cite{gong2019memorizing,zong2018deep}, we designate the ``attack" samples in KDDCUP99 10 percent dataset \cite{Dua:2019} as normal data due to the abundance of ``attack" samples compared to ``non-attack" data. To create our dataset, 50\% of the randomly selected normal data is used for training, while the remaining is reserved for test data. Additionally, 50\% of the anomalous (``non-attack") data is allocated for the test set.

\subsection{Experiment setup}
\label{subsec:experiment_setup}
\subsubsection{Evaluation criteria}
\paragraph{Video datasets} We adhere to the widely popular frame-level area under the ROC curve (AUC) metric \cite{zaheer2020old} for Ped2, Avenue, and ShanghaiTech datasets. 
A higher AUC value indicates more accurate results. Unless otherwise specified, we report the maximum AUC achieved over five repeated experiments.

\paragraph{Image dataset} 
Given an experiment setup in which one image category is selected as normal and the others as anomalous class, AUC is calculated. The setup is repeated for each of all categories as normal, then the AUC values are averaged for the final result.

\paragraph{Network intrusion dataset} Following the protocol in \cite{gong2019memorizing,zong2018deep}, F1, precision, and recall averaged out of 20 repeated runs are reported. Anomaly class is considered as positive class.

\begin{table*}[]
\setlength\tabcolsep{4pt} 
\caption{Architectures used in our video datasets experiments. Each number in the tuple represents time, height, and width dimensions, respectively.}
  \begin{subtable}[t]{0.485\linewidth}
    \caption{$\mathcal{F}$ architecture. Also used in \cite{astrid2021learning}.}
    \resizebox{\linewidth}{!}{%
      \begin{tabular}{|c|l|c|c|c|c|c|}
\hline
\multicolumn{1}{|l|}{}    & Layer           & Out Channels & Filter & Stride    & Padding   & Neg. Slope \\ \hline
\parbox[t]{2mm}{\multirow{12}{*}{\rotatebox[origin=c]{90}{Encoder}}} 
                          & Conv3D          & 96             & (3, 3, 3)   & (1, 2, 2) & (1, 1, 1) & -              \\
                          & BatchNorm3D     & -              & -           & -         & -         & -              \\
                          & LeakyReLU       & -              & -           & -         & -         & 0.2            \\
                          & Conv3D          & 128            & (3, 3, 3)   & (2, 2, 2) & (1, 1, 1) & -              \\
                          & BatchNorm3D     & -              & -           & -         & -         & -              \\
                          & LeakyReLU       & -              & -           & -         & -         & 0.2            \\
                          & Conv3D          & 256            & (3, 3, 3)   & (2, 2, 2) & (1, 1, 1) & -              \\
                          & BatchNorm3D     & -              & -           & -         & -         & -              \\
                          & LeakyReLU       & -              & -           & -         & -         & 0.2            \\
                          & Conv3D          & 256            & (3, 3, 3)   & (2, 2, 2) & (1, 1, 1) & -              \\
                          & BatchNorm3D     & -              & -           & -         & -         & -              \\
                          & LeakyReLU       & -              & -           & -         & -         & 0.2            \\ \hline
\parbox[t]{2mm}{\multirow{11}{*}{\rotatebox[origin=c]{90}{Decoder}}}  
                          & ConvTranspose3D & 256            & (3, 3, 3)   & (2, 2, 2) & (1, 1, 1) & -              \\
                          & BatchNorm3D     & -              & -           & -         & -         & -              \\
                          & LeakyReLU       & -              & -           & -         & -         & 0.2            \\
                          & ConvTranspose3D & 128            & (3, 3, 3)   & (2, 2, 2) & (1, 1, 1) & -              \\
                          & BatchNorm3D     & -              & -           & -         & -         & -              \\
                          & LeakyReLU       & -              & -           & -         & -         & 0.2            \\
                          & ConvTranspose3D & 96             & (3, 3, 3)   & (2, 2, 2) & (1, 1, 1) & -              \\
                          & BatchNorm3D     & -              & -           & -         & -         & -              \\
                          & LeakyReLU       & -              & -           & -         & -         & 0.2            \\
                          & ConvTranspose3D & 1              & (3, 3, 3)   & (1, 2, 2) & (1, 1, 1) & -              \\
                          & Tanh            & -              & -           & -         & -         & -              \\ \hline

\end{tabular}
   }%
  \end{subtable}%
  \hfill 
  \begin{subtable}[t]{0.485\linewidth}
    \caption{$\mathcal{G}$ architecture.}
  \resizebox{\linewidth}{!}{%
      \begin{tabular}{|c|l|c|c|c|c|c|}
\hline
\multicolumn{1}{|l|}{}    & Layer           & Out Channels & Filter & Stride    & Padding   & Neg. Slope \\ \hline
\parbox[t]{2mm}{\multirow{9}{*}{\rotatebox[origin=c]{90}{Encoder}}} 
                          & Conv3D          & 96             & (3, 3, 3)   & (1, 2, 2) & (1, 1, 1) & -              \\
                          & BatchNorm3D     & -              & -           & -         & -         & -              \\
                          & LeakyReLU       & -              & -           & -         & -         & 0.2            \\
                          & Conv3D          & 128            & (3, 3, 3)   & (2, 2, 2) & (1, 1, 1) & -              \\
                          & BatchNorm3D     & -              & -           & -         & -         & -              \\
                          & LeakyReLU       & -              & -           & -         & -         & 0.2            \\
                          & Conv3D          & 256            & (3, 3, 3)   & (2, 2, 2) & (1, 1, 1) & -              \\
                          & BatchNorm3D     & -              & -           & -         & -         & -              \\
                          & LeakyReLU       & -              & -           & -         & -         & 0.2            \\ \hline
\parbox[t]{2mm}{\multirow{8}{*}{\rotatebox[origin=c]{90}{Decoder}}}  
                          & ConvTranspose3D & 128            & (3, 3, 3)   & (2, 2, 2) & (1, 1, 1) & -              \\
                          & BatchNorm3D     & -              & -           & -         & -         & -              \\
                          & LeakyReLU       & -              & -           & -         & -         & 0.2            \\
                          & ConvTranspose3D & 96             & (3, 3, 3)   & (2, 2, 2) & (1, 1, 1) & -              \\
                          & BatchNorm3D     & -              & -           & -         & -         & -              \\
                          & LeakyReLU       & -              & -           & -         & -         & 0.2            \\
                          & ConvTranspose3D & 1              & (3, 3, 3)   & (1, 2, 2) & (1, 1, 1) & -              \\
                          & Tanh * 2           & -              & -           & -         & -         & -              \\ \hline

\end{tabular}
    }%
  \end{subtable}
\label{tab:architecture_video}
\end{table*}

\subsubsection{Input preprocessing}
\paragraph{Video datasets} Following \cite{gong2019memorizing,astrid2021learning}, all frames are converted to grayscale images and then resized to $256 \times 256$. The pixel values ranging from 0 to 255 are normalized into a range from -1 to 1. We take a segment of 16 frames as an input to our model. Therefore, each input is of size $16 \times 1 \times 256 \times 256$, each corresponds to the number of frames, channel size, height, and width.  

\paragraph{Image dataset} Images are resized and normalized from a range from 0 to 255 into a range from 0 to 1. Finally, each input image has a size of $3 \times 32 \times 32$ respectively corresponding to the channel size, height, and width.

\paragraph{Network intrusion dataset} We preprocess each data into features of size 118 dimensions. 
All of the features, both in the training and test sets, are normalized with the minimum and maximum feature values of the training set. 
Consequently, the feature values in the training set has a range of 0 to 1 while the test feature values may not be limited to this range.

\subsubsection{Hyperparameters and implementation details} 
\label{subsec:hyperparameters_and_implementationdetails}
On all the datasets, we train both $\mathcal{F}$ and $\mathcal{G}$ using Adam optimizer \cite{kingma2014adam}. The learning rate of $\mathcal{G}$ is set to $10^{-4}$. For other hyperparameter values, refer to Table \ref{tab:default_hyperparameters}. 
We also provide hyperparameter robustness analysis in Section \ref{subsec:hyperparameter_evaluation}. 

Furthermore, to ensure that $X^P$ (generated pseudo anomly) remains consistent with the input value range of $\mathcal{F}$, e.g., $[-1, 1]$ in video datasets and $[0, 1]$ in image dataset,
we clip $X^P$ (Eq. \eqref{eq:pseudoanomaly}) into the same range.
To integrate the clipped values into the noise amplitude in Eq. \eqref{eq:learningnoise}, $\Delta X$ is recalculated as $X^P - X^N$.


\begin{table*}[]
\setlength\tabcolsep{4pt} 
\caption{Architectures used in our CIFAR10 experiments.}
  \begin{subtable}[t]{0.485\linewidth}
    \caption{$\mathcal{F}$ architecture.}
    \resizebox{\linewidth}{!}{%
      \begin{tabular}{|c|l|c|c|c|c|}
\hline
\multicolumn{1}{|l|}{}    & Layer           & Out Channels & Filter & Stride    & Padding    \\ \hline
\parbox[t]{2mm}{\multirow{12}{*}{\rotatebox[origin=c]{90}{Encoder}}} 
                          & Conv2D          & 64             & 3   & 2 & 0               \\
                          & BatchNorm2D     & -              & -           & -         & -                       \\
                          & ReLU       & -              & -           & -         & -                     \\
                          & Conv2D          & 128            & 3   & 2 & 0               \\
                          & BatchNorm2D     & -              & -           & -         & -                      \\
                          & ReLU       & -              & -           & -         & -                     \\
                          & Conv2D          & 128            & 3   & 2 & 0              \\
                          & BatchNorm2D     & -              & -           & -         & -                       \\
                          & ReLU       & -              & -           & -         & -                     \\
                          & Conv2D          & 256            & 3   & 2 & 0             \\
                          & BatchNorm2D     & -              & -           & -         & -                       \\
                          & ReLU       & -              & -           & -         & -                     \\ \hline
\parbox[t]{2mm}{\multirow{11}{*}{\rotatebox[origin=c]{90}{Decoder}}}  
                          & ConvTranspose2D & 256            & 3   & 2 & 0               \\
                          & BatchNorm2D     & -              & -           & -         & -                     \\
                          & ReLU       & -              & -           & -         & -                  \\
                          & ConvTranspose2D & 128            & 3   & 2 & 0             \\
                          & BatchNorm2D     & -              & -           & -         & -                      \\
                          & ReLU       & -              & -           & -         & -                     \\
                          & ConvTranspose2D & 128             & 3   & 2 & 0              \\
                          & BatchNorm2D     & -              & -           & -         & -                      \\
                          & ReLU       & -              & -           & -         & -                   \\
                          & ConvTranspose2D & 3              & 4   & 2 & 0               \\
                          & Sigmoid           & -              & -           & -         & -                       \\ \hline
\end{tabular}
   }%
  \end{subtable}%
  \hfill 
  \begin{subtable}[t]{0.485\linewidth}
    \caption{$\mathcal{G}$ architecture.}
  \resizebox{\linewidth}{!}{%
      \begin{tabular}{|c|l|c|c|c|c|}
\hline
\multicolumn{1}{|l|}{}    & Layer           & Out Channels & Filter & Stride    & Padding    \\ \hline
\parbox[t]{2mm}{\multirow{9}{*}{\rotatebox[origin=c]{90}{Encoder}}} 
                          & Conv2D          & 64             & 3   & 2 & 0               \\
                          & BatchNorm2D     & -              & -           & -         & -                       \\
                          & ReLU       & -              & -           & -         & -                     \\
                          & Conv2D          & 128            & 3   & 2 & 0               \\
                          & BatchNorm2D     & -              & -           & -         & -                      \\
                          & ReLU       & -              & -           & -         & -                     \\
                          & Conv2D          & 128            & 3   & 2 & 0              \\
                          & BatchNorm2D     & -              & -           & -         & -                       \\
                          & ReLU       & -              & -           & -         & -                     \\                   \hline
\parbox[t]{2mm}{\multirow{8}{*}{\rotatebox[origin=c]{90}{Decoder}}}  
                          
                          & ConvTranspose2D & 128            & 3   & 2 & 0             \\
                          & BatchNorm2D     & -              & -           & -         & -                      \\
                          & ReLU       & -              & -           & -         & -                     \\
                          & ConvTranspose2D & 128             & 3   & 2 & 0              \\
                          & BatchNorm2D     & -              & -           & -         & -                      \\
                          & ReLU       & -              & -           & -         & -                   \\
                          & ConvTranspose2D & 3              & 4   & 2 & 0               \\
                          & Tanh           & -              & -           & -         & -                       \\ \hline

\end{tabular}
    }%
  \end{subtable}
\label{tab:architecture_image}
\end{table*}

\begin{table*}[]
\setlength\tabcolsep{4pt} 
\caption{Architectures used in our KDDCUP experiments.}
  \begin{subtable}[t]{0.4\linewidth}
    \caption{$\mathcal{F}$ architecture. Similar to \cite{gong2019memorizing}.}
    \resizebox{\linewidth}{!}{%
\begin{tabular}{|l|l|c||llc}
\hline
        & Layer  & Out Channels & \multicolumn{1}{l}{}        & \multicolumn{1}{|l|}{Layer}  & \multicolumn{1}{c|}{Out Channels} \\ \hline
\parbox[t]{2mm}{\multirow{8}{*}{\rotatebox[origin=c]{90}{Encoder}}} & Linear & 60           & \parbox[t]{2mm}{\multirow{7}{*}{\rotatebox[origin=c]{90}{Decoder}}} & \multicolumn{1}{|l|}{Linear} & \multicolumn{1}{c|}{10}           \\
        & Tanh   & -            & \multicolumn{1}{l}{}        & \multicolumn{1}{|l|}{Tanh}   & \multicolumn{1}{c|}{-}            \\
        & Linear & 30           & \multicolumn{1}{l}{}        & \multicolumn{1}{|l|}{Linear} & \multicolumn{1}{c|}{30}           \\
        & Tanh   & -            & \multicolumn{1}{l}{}        & \multicolumn{1}{|l|}{Tanh}   & \multicolumn{1}{c|}{-}            \\
        & Linear & 10           & \multicolumn{1}{l}{}        & \multicolumn{1}{|l|}{Linear} & \multicolumn{1}{c|}{60}           \\
        & Tanh   & -            & \multicolumn{1}{l}{}        & \multicolumn{1}{|l|}{Tanh}   & \multicolumn{1}{c|}{-}            \\
        & Linear & 3            & \multicolumn{1}{l}{}        & \multicolumn{1}{|l|}{Linear} & \multicolumn{1}{c|}{118}          \\ \hhline{|~|~|~|---} 
        & Tanh   & -            &                              &                             & \multicolumn{1}{l}{}              \\ \hhline{|---|~~~} 
\end{tabular}

   }%
  \end{subtable}%
  \hfill 
  \begin{subtable}[t]{0.4\linewidth}
    \caption{$\mathcal{G}$ architecture.}
  \resizebox{\linewidth}{!}{%
      \begin{tabular}{|l|l|c||llc}
\hline
        & Layer  & Out Channels & \multicolumn{1}{l}{}        & \multicolumn{1}{|l|}{Layer}  & \multicolumn{1}{c|}{Out Channels} \\ \hline
\parbox[t]{2mm}{\multirow{6}{*}{\rotatebox[origin=c]{90}{Encoder}}} & Linear & 60           & \parbox[t]{2mm}{\multirow{5}{*}{\rotatebox[origin=c]{90}{Decoder}}} & \multicolumn{1}{|l|}{Linear} & \multicolumn{1}{c|}{30}           \\
        & Tanh   & -            & \multicolumn{1}{l}{}        & \multicolumn{1}{|l|}{Tanh}   & \multicolumn{1}{c|}{-}            \\
        & Linear & 30           & \multicolumn{1}{l}{}        & \multicolumn{1}{|l|}{Linear} & \multicolumn{1}{c|}{60}           \\
        & Tanh   & -            & \multicolumn{1}{l}{}        & \multicolumn{1}{|l|}{Tanh}   & \multicolumn{1}{c|}{-}            \\
        & Linear & 10           & \multicolumn{1}{l}{}        & \multicolumn{1}{|l|}{Linear} & \multicolumn{1}{c|}{118}           \\ \hhline{|~|~|~|---} 
        & Tanh   & -            & \multicolumn{1}{l}{}        &     &           \\          \hhline{|---|~~~} 
\end{tabular}
}%
  \end{subtable}

\label{tab:architecture_network}
\end{table*}

\subsubsection{Architectures}
\label{subsec:architectures}
The neural network architectures used on the video, image, and network intrusion datasets are provided in Table \ref{tab:architecture_video}, \ref{tab:architecture_image}, and \ref{tab:architecture_network}, respectively. Encoder of $\mathcal{F}$ produces a latent code of size $T \times C \times H \times W = 2 \times 256 \times 16 \times 16$ for video datasets, $C \times H \times W = 256 \times 1 \times 1$ for image dataset, and $C = 3$ for network intrusion dataset.

The final layer of $\mathcal{F}$ aligns with the input range. Specifically, a tanh final layer is applied for video data within the range of $[-1, 1]$ (Table \ref{tab:architecture_video}(a)), a sigmoid final layer for image data within the range of $[0, 1]$ (Table \ref{tab:architecture_image}(a)), and a linear final layer for network intrusion dataset (Table \ref{tab:architecture_network}(a)). It may also be noted that for bounded data (i.e., video and image), the final activation function of $\mathcal{G}$ is twice of the corresponding output layer activation of $\mathcal{F}$ (Table \ref{tab:architecture_video}(b), \ref{tab:architecture_image}(b)). For instance, tanh * 2 of $\mathcal{G}$ for tanh of $\mathcal{F}$. This configuration allows the generated noise to modify an input from one extreme value to the other, enabling, for example, a pixel with a value $-1$ to be changed to $+1$ by a noise value of $+2$. 


\begin{figure*}[t]
\begin{center}
\includegraphics[width=\linewidth]{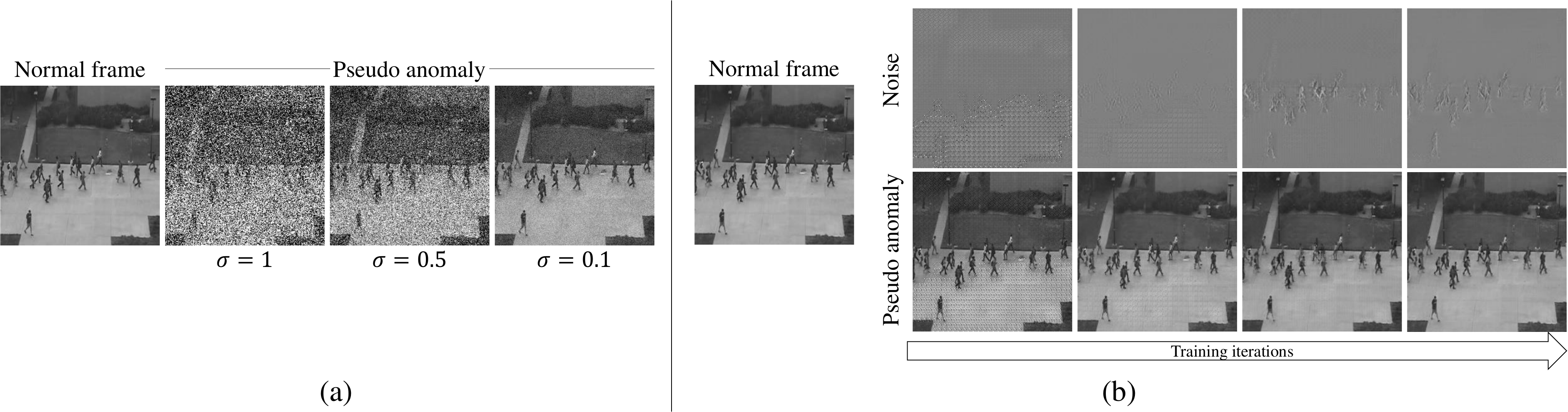}
\end{center}
   \caption{Visualizations of (a) pseudo anomalies constructed from a normal frame by adding Gaussian noise with various $\sigma$ values, where the random noise amplitude is affected by $\sigma$; and (b) noise generated by $\mathcal{G}$ and the respective pseudo anomalies generated using our proposed \textit{learning to generate pseudo anomalies} mechanism across different training iterations. Compared to the random noise in (a), the noise generated in our propose mechanism changes with training iterations as $\mathcal{G}$ adapts to the reconstruction boundary of $\mathcal{F}$.}
\label{fig:pseudoanomaly}
\end{figure*}

\subsubsection{Anomaly score calculation} 
\label{subsubsec:anomalyscorecalculation}

\paragraph{Video datasets}

Concurrent to \cite{park2020learning,liu2018future,astrid2021learning,astrid2021synthetic}, for a given test video comprising frames $\{X_1, X_2, ..., X_n\}$, we feed $\mathcal{F}$ with a 16-frame input sequence $\{X_t, X_{t+1}, ..., X_{t+15}\}$, for $t=1$ to $t=n-15$. We then compute the Peak Signal-to-Noise Ratio (PSNR) value $\mathcal{P}_{t+8}$ between the 9th frame of the input $X_{t+8}$ and its reconstruction $\hat{X}_{t+8}$:
\begin{equation}
    \mathcal{P}_{t+8} = 10 \text{ log}_{10}  \frac{M_{\hat{X}_{t+8}}^2}{\frac{1}{R} \left \| \hat{X}_{t+8} - X_{t+8}  \right \|_{F}^{2} } \text{,}
\label{eq:psnr}
\end{equation}
where $R$ represents the total number of pixels in $\hat{X}_{t+8}$, and $M_{\hat{X}_{t+8}}$ is the maximum possible pixel value of $\hat{X}_{t+8}$, i.e., $M_{\hat{X}_{t+8}} = 1$ since the video frames are normalized to $[-1, 1]$. Min-max normalization is applied on the PSNR values across $t=1$ to $t=n-15$ of a test video to obtain the normalcy score $\mathcal{Q}_{t+8}$ within the range of $[0,1]$.
Finally, we calculate the anomaly score $\mathcal{A}_{t+8}$ for each input sequence as:
\begin{equation}
    \mathcal{A}_{t+8} = 1 - \mathcal{Q}_{t+8} \text{.}
\label{eq:anomalyscore}
\end{equation}

\begin{table*}[]
\caption{Ablation studies comparing $\mathcal{F}$ trained without pseudo anomalies, $\mathcal{F}$ trained with non-learnable Gaussian noise, and $\mathcal{F}$ trained with learnable noise. The best performances are marked as bold.}
\small
\centering
\resizebox{\linewidth}{!}{
\begin{tabular}{|l||c|c|c||c||ccc|}
\hline
\multirow{2}{*}{Pseudo anomaly} & Ped2  & Avenue & ShanghaiTech & CIFAR-10 & \multicolumn{3}{c|}{KDDCUP} \\
                                & AUC   & AUC    & AUC          & AUC      & F1     & Precision & Recall \\ \hline
None (baseline)                 & 92.49 & 81.47  & 71.28        & 60.10    & 94.53  & 93.94     & 95.13  \\
Gaussian noise $\sigma=0.1$              & 93.32 & 81.56  & 71.24        & 63.20    & 94.52  & 93.78     & 95.28  \\
Gaussian noise $\sigma=0.5$               & 93.12 & 82.10   & 71.73        & 61.80    & 94.78  & 94.04     & 95.53  \\
Gaussian noise $\sigma=1$                 & 93.03 & 82.09  & 71.92        & 61.91    & 95.52  & 94.78     & 96.27  \\
Learnable noise (ours)          & \textbf{94.57} & \textbf{83.23}  & \textbf{73.23}        & \textbf{67.76}    & \textbf{95.58}  & \textbf{94.84}     & \textbf{96.34}  \\ \hline
\end{tabular}
}
\label{tab:ablation}
\end{table*}

\paragraph{Image dataset}
Given a test input image $X$ and its reconstruction $\hat{X}$, we calculate the anomaly score using the reconstruction loss as:
\begin{equation}
    \mathcal{A} = \left \| \hat{X} - X  \right \|_{F}^{2} \text{.}
\label{eq:anomalyscoreimageandkdd}
\end{equation}
We then min-max normalize $\mathcal{A}$ across the test data to get scores ranging from $0$ to $1$.

\paragraph{Network intrusion dataset}
We first calculate the anomaly score $\mathcal{A}$ similarly to the calculation on image dataset. However, as the evaluation metrics (i.e., F1, precision, recall) on the network intrusion dataset necessitate a threshold to distinguish normal from anomaly, unlike AUC for the image dataset, we select the predictions with the top 20\% anomaly scores across the test set as anomaly/positive, while the rest as normal/negative, following the procedure outlined in \cite{zong2018deep}.

\subsection{Ablation studies}
\label{subsec:ablation}

To evaluate the effectiveness of utilizing learnable noise to generate pseudo anomalies, 
we compare our method with a model wherein the main autoencoder $\mathcal{F}$ is trained solely on normal data (baseline) and another model wherein $\mathcal{F}$ is trained with non-learnable noise-based pseudo anomalies. Training using only normal data is equivalent to setting the probability $p$ to $0$.
For experiments involving non-learnable noise-based pseudo anomalies, we generate these pseudo anomalies by adding non-learnable Gaussian noise augmentation to the normal data $X^N$. 
The noise at the $i$-th position $\Delta X_i$ (Eq. \ref{eq:pseudoanomaly}) is defined as: 
\begin{equation}
    \Delta X_i = \mathcal{N}(0, \sigma) \text{,}
\label{eq:pseudogaussian}
\end{equation}
where $\mathcal{N}(0, \sigma)$ is Gaussian noise with mean $0$ and standard deviation $\sigma$. Examples of pseudo anomalies generated with different $\sigma$ values can be seen in Figure \ref{fig:pseudoanomaly}(a). 

\begin{figure*}[t]
\begin{center}
\includegraphics[width=\linewidth]{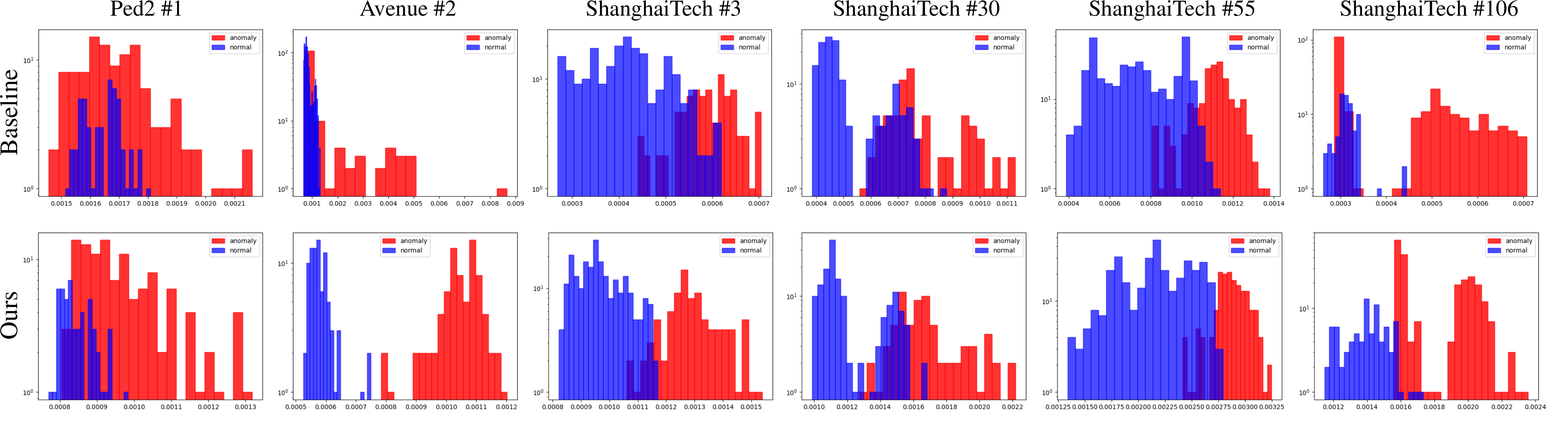}
\end{center}
   \caption{The distribution of reconstruction errors of the baseline and our model for normal (blue) and anomalous (red) data in several videos. It is evident that the reconstruction error distribution becomes more discriminative with our model compared to the baseline.}
\label{fig:recon_loss_hist}
\end{figure*}

\begin{table*}[]
\centering
\caption{Mean anomaly scores of anomalous data ($\mu^\mathcal{A}_A$) and normal data ($\mu^\mathcal{A}_N$). A higher difference $\mu^\mathcal{A}_A - \mu^\mathcal{A}_N$ (highlighted in bold) indicates better discrimination between anomalous and normal data.}
\resizebox{0.7\linewidth}{!}{
\begin{tabular}{|l|ccc|ccc|}
\hline
                          & \multicolumn{3}{c|}{Baseline}                                                                       & \multicolumn{3}{c|}{Ours}                                                                                   \\ 
\multirow{-2}{*}{Dataset} & \multicolumn{1}{c|}{$\mu^\mathcal{A}_A$} & \multicolumn{1}{c|}{$\mu^\mathcal{A}_N$} &\cellcolor[HTML]{D6D6D6}{$\mu^\mathcal{A}_A - \mu^\mathcal{A}_N$}    & \multicolumn{1}{c|}{$\mu^\mathcal{A}_A$} &\multicolumn{1}{c|}{$\mu^\mathcal{A}_N$} & \cellcolor[HTML]{D6D6D6}{$\mu^\mathcal{A}_A - \mu^\mathcal{A}_N$}             \\ \hline
Ped2                      & \multicolumn{1}{c|}{0.6150}      & \multicolumn{1}{c|}{0.1929}     & \cellcolor[HTML]{D6D6D6}0.4220 & \multicolumn{1}{c|}{0.6169}      & \multicolumn{1}{c|}{0.1620}     & \cellcolor[HTML]{D6D6D6}\textbf{0.4549} \\
Avenue                    & \multicolumn{1}{c|}{0.5122}      & \multicolumn{1}{c|}{0.2351}     & \cellcolor[HTML]{D6D6D6}0.2771 & \multicolumn{1}{c|}{0.5230}      & \multicolumn{1}{c|}{0.2345}     & \cellcolor[HTML]{D6D6D6}\textbf{0.2885} \\
ShanghaiTech              & \multicolumn{1}{c|}{0.5686}      & \multicolumn{1}{c|}{0.3763}     & \cellcolor[HTML]{D6D6D6}0.1923 & \multicolumn{1}{c|}{0.5692}      & \multicolumn{1}{c|}{0.3753}     & \cellcolor[HTML]{D6D6D6}\textbf{0.1939} \\
CIFAR-10                  & \multicolumn{1}{c|}{0.7416}      & \multicolumn{1}{c|}{0.7133}     & \cellcolor[HTML]{D6D6D6}0.0282 & \multicolumn{1}{c|}{0.7765}      & \multicolumn{1}{c|}{0.7125}     & \cellcolor[HTML]{D6D6D6}\textbf{0.0640} \\
KDDCUP                    & \multicolumn{1}{c|}{0.9998}      & \multicolumn{1}{c|}{0.9913}     & \cellcolor[HTML]{D6D6D6}0.0084 & \multicolumn{1}{c|}{0.9997}      & \multicolumn{1}{c|}{0.9866}     & \cellcolor[HTML]{D6D6D6}\textbf{0.0131} \\ \hline
\end{tabular}
}
\label{tab:anomaly_score}
\end{table*}


The comparisons between $\mathcal{F}$ trained without pseudo anomalies, $\mathcal{F}$ trained with Gaussian noise with various $\sigma$ values, and $\mathcal{F}$ trained with our learnable noise mechanism can be seen in Table \ref{tab:ablation}. As seen, adding Gaussian noise to create pseudo anomalies can generally improve the model's performance, highlighting the importance of noise-based pseudo anomalies. However, using learnable noise can further enhance the model's performances which shows the significance of our proposed learning component. 
We can also observe in Figure \ref{fig:pseudoanomaly}(b) that noise generation progresses as the training goes. This learning contributes to the performance gain and high discrimination capability for our anomaly detection model that utilizes learnable noise over the non-learnable Gaussian noise. 

To gain further insights into how our pseudo anomalies affect the reconstruction capability of the model on normal and anomalous data, Figure \ref{fig:recon_loss_hist} displays the distribution of reconstruction errors on several videos. 
As observed, our model exhibits a more prominent separation between the normal and anomalous data distributions compared to the baseline, indicating a better discrimination capability.
It is worth noting that what is important is the reconstruction error distribution difference between the normal and anomalous data, rather than the magnitude of the reconstruction errors.
Similarly, as for the anomaly score reported in Table \ref{tab:anomaly_score}, our model observes a higher difference between anomalous and normal data scores indicating a better discrimination capability.

\subsection{Comparisons with SOTA}
\label{subsec:comparisons_sota}
To evaluate the performance and generic applicability of our method, in Table \ref{tab:sota_all}, we compare our approach with other state-of-the-arts (SOTA) methods on video datasets (i.e., Ped2, Avenue, and ShanghaiTech), an image dataset (i.e., CIFAR-10), and a network intrusion dataset (i.e., KDDCUP). Following \cite{gong2019memorizing}, we categorize the methods into Reconstruction and Non-Reconstruction. Our method belongs to reconstruction-based methods which use reconstruction quality to measure anomaly scores.

While most SOTA methods demonstrate superior performance on video datasets, they often introduce inductive biases, such as assuming movement during anomalous events or the presence of anomalous objects. Consequently, these methods are not applicable to all three types of task (see results with `$\times$' marks in Table \ref{tab:sota_all}). Specifically, skip frame-based methods \cite{astrid2021learning,astrid2021synthetic,astrid2023pseudobound,georgescu2021anomaly}, methods altering the order of frames \cite{astrid2022limiting,georgescu2021anomaly}, those strictly employing Recurrent Neural Network (RNN), Long Short-Term Memory (LSTM), or 3D-convolution \cite{lee2019bman,zhao2017spatio,luo2017remembering,luo2017revisit}, prediction-based methods predicting missing frames \cite{georgescu2021anomaly,lu2020few,liu2018future,park2020learning,zhao2017spatio}, and methods requiring optical flow \cite{wang2023memory,vu2019robust,georgescu2021background,liu2018future,hasan2016learning} are incompatible with data without movements, such as image and network intrusion data. 
Furthermore, methods relying on an object detector \cite{georgescu2021background,georgescu2021anomaly,sun2023hierarchical} are dependent on object properties available only in video datasets.
Patch-based methods, which inject patches from another dataset to synthesize pseudo anomalies \cite{astrid2021learning,astrid2021synthetic} may work in both image and video domains but are not suitable for the network intrusion dataset. Additionally, their performances on image datasets remain undetermined as the results have not been reported.

Several others that do not impose the aforementioned limitations, such as OGNet \cite{zaheer2020old}, Pseudobound-Noise \cite{astrid2023pseudobound}, and MNAD-Reconstruction \cite{park2020learning}, theoretically can be generic and work with non-video domains, however nothing has been reported about their performance on non-video datasets. 
One exceptional SOTA model is a memory-based method, MemAE \cite{gong2019memorizing}, that is evaluated on all video, image, and network intrusion datasets. In comparison, our method outperforms MemAE on video and image datasets while competitive on the KDDCUP network intrusion dataset. This underscores the generic applicability and superiority of our approach compared to other SOTA methods.

\begin{table*}[t]
\caption{Performance comparisons of our approach and the existing state-of-the-art methods on video datasets (i.e., Ped2 \cite{li2013anomaly}, Avenue \cite{lu2013abnormal}, and ShanghaiTech \cite{luo2017revisit}), image dataset (CIFAR-10 \cite{krizhevsky2009learning}), and network intrusion dataset (i.e., KDDCUP \cite{Dua:2019}). Best and second best in each category and dataset are marked as bold and underlined. Results that are impossible to obtain due to inductive bias are marked as `$\times$', whereas results that are not reported in the original paper but in principle possible to get are marked as `-'.}
\resizebox{\linewidth}{!}{
\centering
\begin{tabular}{cl||c|c|c||c||ccc|}
\hline
\multicolumn{2}{c||}{\multirow{2}{*}{Methods}}                                         & Ped2  & Avenue & ShanghaiTech & CIFAR-10 & \multicolumn{3}{c|}{KDDCUP} \\
\multicolumn{2}{c||}{}                                                                 & AUC   & AUC    & AUC          & AUC      & F1     & Precision & Recall \\ \hline
\multicolumn{1}{c|}{\parbox[t]{2mm}{\multirow{10}{*}{\rotatebox[origin=c]{90}{Non-Reconstruction}}}} & MAAM-Net \cite{wang2023memory}                & 97.7  & 90.9   & 71.3         & $\times$        & $\times$       & $\times$          & $\times$       \\
\multicolumn{1}{c|}{}                                    & BMAN \cite{lee2019bman}                        & 96.6  & 90.0   & 76.2         & $\times$         & $\times$       & $\times$          & $\times$       \\
\multicolumn{1}{c|}{}                                    & Vu \etal \cite{vu2019robust}         & \underline{99.21} & 71.54  & -            & $\times$         & $\times$       & $\times$          & $\times$       \\
\multicolumn{1}{c|}{}                                    & OGNet \cite{zaheer2020old}                  & 98.1  & -      & -            & -        & -      & -         & -      \\
\multicolumn{1}{c|}{}                                    & Georgescu \etal \cite{georgescu2021background}  & 98.7  & 92.3   & 82.7         & $\times$         & $\times$       & $\times$          & $\times$       \\
\multicolumn{1}{c|}{}                                    & Georgescu \etal \cite{georgescu2021anomaly}  & \textbf{99.8}  & \underline{92.8}   & \textbf{90.2}         & $\times$         & $\times$       & $\times$          & $\times$       \\
\multicolumn{1}{c|}{}                                    & HSC \cite{sun2023hierarchical}                        & 98.1  & \textbf{93.7}   & \underline{83.4}         & $\times$         & $\times$       & $\times$          & $\times$       \\
\multicolumn{1}{c|}{} & Frame-Pred  \cite{liu2018future}          & 95.4  & 85.1  & 72.8 
 & $\times$         & $\times$       & $\times$          & $\times$       \\
\multicolumn{1}{c|}{} & Lu \etal \cite{lu2020few}         & 96.2  & 85.8   & 77.9         & $\times$         & $\times$       & $\times$          & $\times$       \\
\multicolumn{1}{c|}{}                                    & MNAD-Prediction \cite{park2020learning}            & 97.0  & 88.5   & 70.5         & $\times$         & $\times$       & $\times$          & $\times$       \\ \hline
\multicolumn{1}{c|}{\parbox[t]{2mm}{\multirow{15}{*}{\rotatebox[origin=c]{90}{Reconstruction}}}}    & AE-Conv2D  \cite{hasan2016learning}                & 90.0  & 70.2   & 60.85        & $\times$         & $\times$       & $\times$          & $\times$       \\
\multicolumn{1}{c|}{}                                    & AE-Conv3D  \cite{zhao2017spatio}                   & 91.2  & 71.1   & -            & $\times$         & $\times$       & $\times$          & $\times$       \\
\multicolumn{1}{c|}{}                                    & AE-ConvLSTM  \cite{luo2017remembering}       & 88.10 & 77.00  & -            & $\times$         & $\times$       & $\times$          & $\times$       \\
\multicolumn{1}{c|}{}                                    & TSC \cite{luo2017revisit}                       & 91.03 & 80.56  & 67.94        & $\times$         & $\times$       & $\times$          & $\times$       \\
\multicolumn{1}{c|}{}                                    & StackRNN \cite{luo2017revisit}                  & 92.21 & 81.71  & 68.00        & $\times$         & $\times$       & $\times$          & $\times$       \\
\multicolumn{1}{c|}{}                                    & MemAE \cite{gong2019memorizing}                      & 94.1  & 83.3   & 71.2         & \underline{60.88}    & \textbf{96.41}  & \textbf{96.27}     & \textbf{96.55}  \\
\multicolumn{1}{c|}{}                                    & MNAD-Reconstruction \cite{park2020learning}       & 90.2  & 82.8   & 69.8         & -        & -      & -         & -      \\
\multicolumn{1}{c|}{}                                    & Astrid \etal \cite{astrid2022limiting}   & 93.46 & 81.78  & -            & $\times$         & $\times$       & $\times$          & $\times$       \\
\multicolumn{1}{c|}{}                                    & PseudoBound-Noise \cite{astrid2023pseudobound} & \underline{97.78} & 82.11  & 72.02        & -        & -      & -         & -      \\
\multicolumn{1}{c|}{}                                    & PseudoBound-Patch \cite{astrid2023pseudobound}  & 95.33 & \underline{85.36}  & 72.77        & -        & $\times$       & $\times$          & $\times$       \\
\multicolumn{1}{c|}{}                                    & STEAL Net \cite{astrid2021synthetic,astrid2023pseudobound}      & \textbf{98.44} & \textbf{87.10}  & \underline{73.66}        & $\times$         & $\times$       & $\times$          & $\times$       \\
\multicolumn{1}{c|}{}                                    & LNTRA-Patch \cite{astrid2021learning}                & 94.77 & 84.91  & 72.46        & -        & $\times$       & $\times$          & $\times$       \\
\multicolumn{1}{c|}{}                                    & LNTRA-Skip frame \cite{astrid2021learning}           & 96.50 & 84.67  & \textbf{75.97}        & $\times$         & $\times$       & $\times$          & $\times$       \\
\cdashline{2-9}
\multicolumn{1}{c|}{}                                    & Baseline                   & 92.49 & 81.47  & 71.28        & 60.10    & 94.53  & 93.94     & 95.13  \\
\multicolumn{1}{c|}{}                                    & Ours                       & 94.57 & 83.23  & 73.23        & \textbf{67.76}    & \underline{95.58}  & \underline{94.84}     & \underline{96.34}  \\ \hline
\end{tabular}
}
\label{tab:sota_all}
\end{table*}

\subsection{Hyperparameter evaluation}
\label{subsec:hyperparameter_evaluation}

In this work, we introduce two new hyperparameters, $p$ and $\lambda$. Here, $p$ represents the probability of using pseudo anomaly (Eq. \eqref{eq:aepseudo}), while $\lambda$ is the weighting factor that controls the amount of noise relative to generating pseudo anomaly within the reconstruction boundary (Eq. \eqref{eq:learningnoise}). This section delves into the effects of hyperparameters values on the model's performance, specifically on Ped2, CIFAR-10, and KDDCUP. On Ped2 and KDDCUP, we repeat each setup five and twenty times respectively and report the average and maximum performance. 
For CIFAR-10, we conduct the experiment once and report the AUC averaged out of the ten normal classes setup (refer to Section \ref{subsec:datasets}).

\subsubsection{Pseudo anomaly ratio $p$}
\begin{figure*}
\begin{center}
\includegraphics[width=\linewidth]{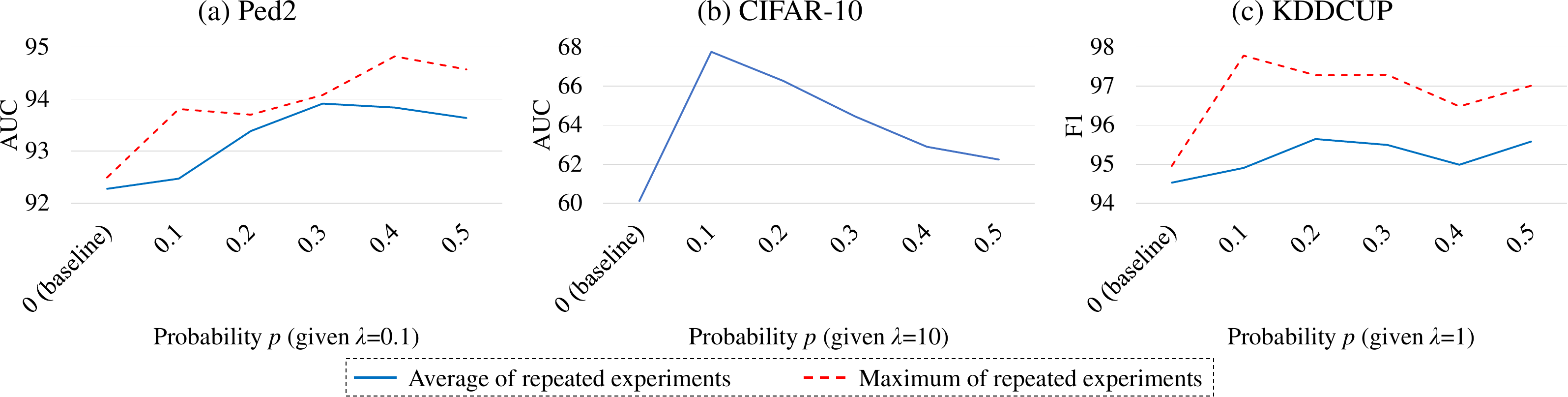}
\end{center}
   \caption{Probability $p$ hyperparameter evaluation on (a) Ped2, (b) CIFAR-10, and (c) KDDCUP. The baseline (training without pseudo anomaly) is equivalent to $p=0$. Our approach is robust across different $p$ values, shown by better performances against the baseline.
   }
\label{fig:hyperparameter_probability}
\end{figure*}

Figure \ref{fig:hyperparameter_probability} shows the results on different values of $p$ on Ped2, CIFAR-10, and KDDCUP. The baseline is equivalent to $p=0$. It can be observed that any $p$ values other than zero outperforms the baseline demonstrating that our model is robust to diverse $p$ values.

\begin{figure*}
\begin{center}
\includegraphics[width=\linewidth]{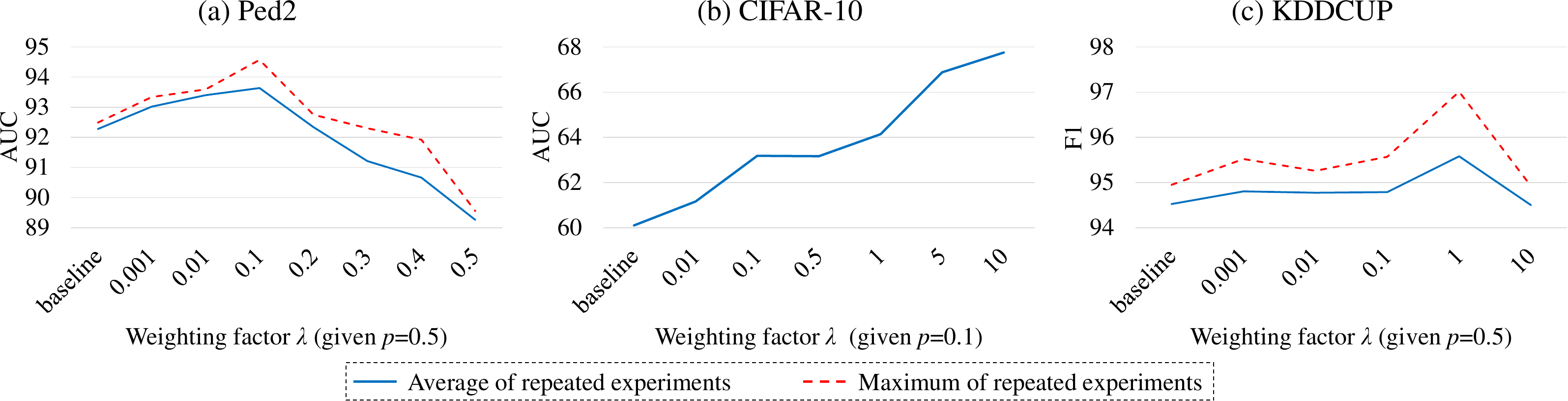}
\end{center}
   \caption{Weighting factor $\lambda$ hyperparameter evaluation on (a) Ped2, (b) CIFAR-10, and (c) KDDCUP. In the case of Ped2 and KDDCUP, too high $\lambda$ values can be harmful to the model. For CIFAR-10, while higher $\lambda$ values are superior, smaller values are also acceptable (i.e., better than the baseline). Therefore, setting a smaller $\lambda$ value is relatively robust to any problems.}
\label{fig:hyperparameter_weighting}
\end{figure*}

\begin{figure*}
\begin{center}
\includegraphics[width=\linewidth]{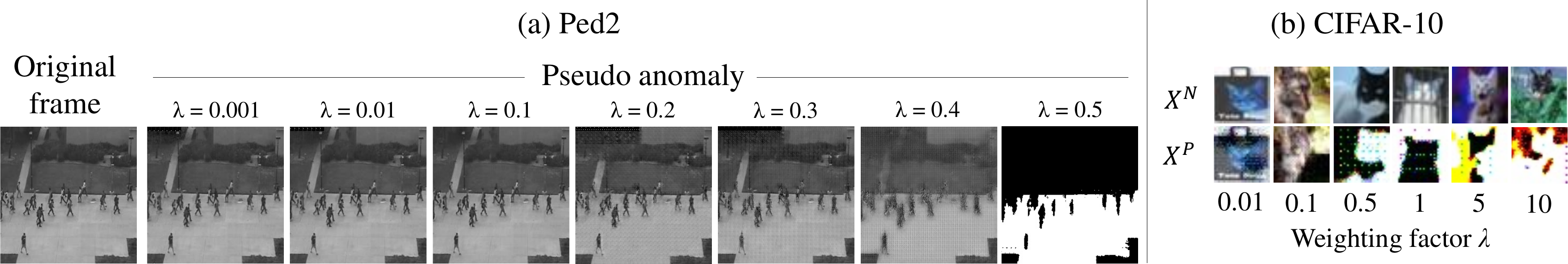}
\end{center}
   \caption{Visualization of pseudo anomalies generated using different $\lambda$ values on (a) Ped2 and (b) CIFAR-10 (`cat' class as the normal class). Higher $\lambda$ values produce pseudo anomalies quite different from the normal data.}
\label{fig:hyperparameter_weighting_qualitative}
\end{figure*}

\begin{figure*}[t]
\begin{center}
\includegraphics[width=\linewidth]{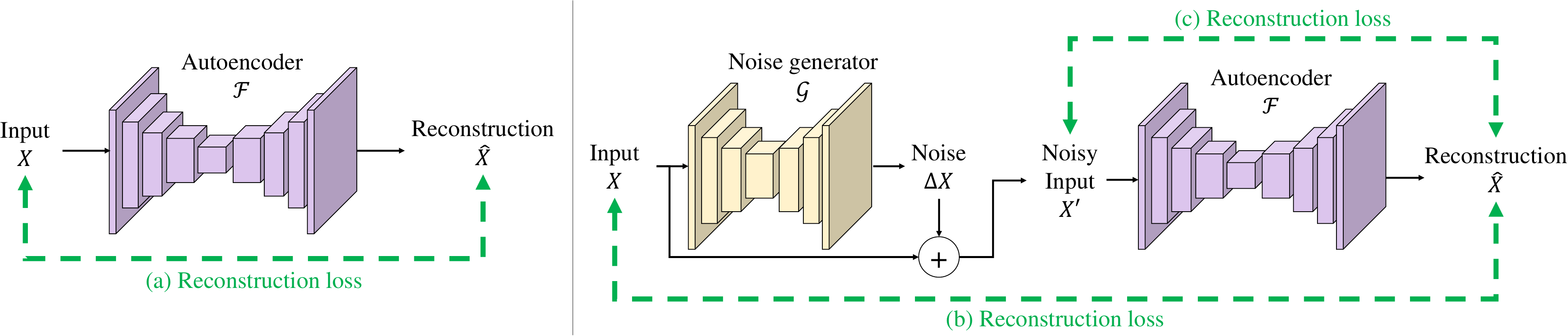}
\end{center}
   \caption{During test time, anomaly scores are obtained from the reconstruction loss, meaning high loss values correspond to high anomaly scores. (a) As our default configuration, we utilize only the trained AE (either with input-level or feature-level pseudo anomalies). When the noise generator $\mathcal{G}$ is incorporated into the test pipeline, then we may have 
   two options in computing the reconstruction loss: (b) with the clean input or (c) noisy input.}
\label{fig:testconfiguration}
\end{figure*}

\subsubsection{Noise weighting factor $\lambda$}

In Eq. \eqref{eq:learningnoise}, a higher noise weighting factor $\lambda$ value gives more importance to noise in relative to generating pseudo anomalies with low reconstruction error of the AE.
Figure \ref{fig:hyperparameter_weighting} shows the results for different values of $\lambda$ on Ped2, CIFAR-10, and KDDCUP. As seen in Figure \ref{fig:hyperparameter_weighting}(a) \& (c), putting too much importance on high noise can be harmful to the model, as observed in our experiments on Ped2 and KDDCUP. Visualization of pseudo anomalies generated by the model at the end of training on Ped2 can be seen in Figure \ref{fig:hyperparameter_weighting_qualitative}(a), which shows that when $\lambda \geq 0.2$, highly noisy pseudo anomalies are generated. However, unlike the video and network intrusion datasets, stronger noise works better in case of CIFAR-10 dataset, as seen in Figure \ref{fig:hyperparameter_weighting}(b). This may be attributed to the characteristics of the image anomalies that are far from normal data distribution, i.e., image anomalies are different image categories from normal ones. 
As such, as visualized in Figure \ref{fig:hyperparameter_weighting_qualitative}(b), highly distorted pseudo anomalies are learned to help the model to detect images of categories different from the normal category. 
Nevertheless, as seen in \ref{fig:hyperparameter_weighting}(b), using smaller noise also works comparably better than the baseline. Therefore, from this experiment, we can conclude that setting smaller $\lambda$ values, e.g., $0 < \lambda \leq 0.1$, are robust across different types of datasets such as video, image, or network intrusion, demonstrating the higher importance of generating pseudo anomalies inside the reconstruction boundary in comparison to generating pseudo anomalies way far away from the normal data for generic applicability. 

\begin{table*}[]
\caption{Comparisons of various test setups. \textit{Without $\mathcal{G}$}, \textit{With $\mathcal{G}$ (loss target: $X$)}, and \textit{With $\mathcal{G}$ (loss target: $X'$)} are illustrated in Figure \ref{fig:testconfiguration}(a), (b), and (c), respectively.  }
\small
\centering
\resizebox{\linewidth}{!}{
\begin{tabular}{|l|c|c|c|c|ccc|}
\hline
\multirow{2}{*}{Test   method}    & Ped2  & Avenue & ShanghaiTech & CIFAR-10 & \multicolumn{3}{c|}{KDDCUP} \\
                                  & AUC   & AUC    & AUC          & AUC      & F1     & Precision & Recall \\ \hline
Without $\mathcal{G}$                 & \textbf{94.57} & \textbf{83.23}  & \textbf{73.22}        & \textbf{67.76}    & \textbf{95.58}  & \textbf{94.84}     & \textbf{96.34}  \\
With $\mathcal{G}$ (loss target: $X$) & 91.47 & 79.99  & 65.05        & 59.69    & 33.70  & 33.11     & 33.63  \\
With $\mathcal{G}$ (loss target: $X'$) & 91.24 & 77.94  & 68.67        & 39.69    & 28.44  & 19.25     & 68.28  \\ \hline
\end{tabular}}
\label{tab:testconfiguration}
\end{table*}

\subsection{Additional discussions}
\label{subsec:additional_discussions}
In this subsection, we discuss our design choice of not using generator during test time (Section \ref{subsec:isgeneratornecessary}) and provide comparison with adversarial denoising AE in the term of training progression (Section \ref{subsec:trainingprogression}).

\subsubsection{Is generator necessary at test time?}
\label{subsec:isgeneratornecessary}
In our default configuration, we use only $\mathcal{F}$ during test time (Figure \ref{fig:testconfiguration}(a)). $\mathcal{G}$ is used only at training to generate pseudo anomalies. However, one may wonder how things will go when the trained $\mathcal{G}$ is incorporated into the test pipeline.
As seen in Figure \ref{fig:testconfiguration}(b) and (c), we put the trained $\mathcal{G}$ and feed the input to $\mathcal{G}$ to create noisy input $X'$ that will be given to $\mathcal{F}$. In this setting, we can have two options on calculating the reconstruction error: with clean input $X$ (Figure \ref{fig:testconfiguration}(b)) or noisy input $X'$ (Figure \ref{fig:testconfiguration}(c)). 

AUC comparisons on Ped2, Avenue, ShanghaiTech, and CIFAR-10 of different test configurations can be seen in Table \ref{tab:testconfiguration}. Testing without $\mathcal{G}$ consistently achieves the highest performances, supporting that the trained $\mathcal{F}$ based on our approach is capable of well-reconstructing normal data while poorly-reconstructing anomalous data. Testing with $\mathcal{G}$ is not only harmful to the model performance but also add more inference time.

\begin{figure*}[h]
\begin{center}
\includegraphics[width=\linewidth]{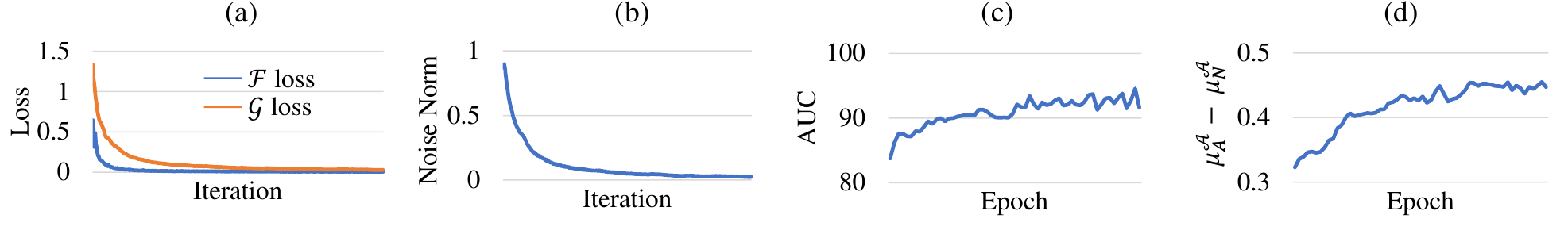}
\end{center}
  \caption{Training progression on Ped2 dataset seen in (a) $\mathcal{F}$ loss (Eq. \eqref{eq:reconloss} \& \eqref{eq:pseudoreconloss}) and $\mathcal{G}$ loss (Eq. \eqref{eq:learningnoise}), (b) generated noise norm, (c) AUC,  and (d) difference of mean anomaly scores between anomalous and normal data ($\mu^\mathcal{A}_A - \mu^\mathcal{A}_N$). Stable progression during the training shows the stability of our approach in enhancing the reconstruction boundary of the AE.}
\label{fig:trend}
\end{figure*}

\subsubsection{Training progression and comparisons with adversarial denoising AE}
\label{subsec:trainingprogression}

Despite the architectural similarity to adversarial denoising autoencoder approaches \cite{salehi2021arae,jewell2022one}, our method is rather a \textit{cooperative} learning between the generator $\mathcal{G}$ and $\mathcal{F}$ (`discriminator'), which can be supported by the mutual loss decrease and convergence (Figure \ref{fig:trend} (a)). 
Additionally, the convergence can also be seen in the AUC trend over the training epochs in Figure \ref{fig:trend}(c). As our method can converge in a cooperative way, it is evidently more stable compared to the adversarial training methods \cite{salehi2021arae,jewell2022one} in which the loss tends to fluctuate due to adversarial training objectives and therefore difficult to converge. Furthermore, there is a hint of instability in \cite{jewell2022one} due to its delicate selection of L1/L2 loss and hyperparameters while not providing any hyperparameter sensitivity evaluation. Moreover, both these methods only report the results on simple datasets and their performance on highly complex datasets such as ShanghaiTech is unknown.

The cooperation between $\mathcal{G}$ and $\mathcal{F}$ throughout training can also be observed in Figure \ref{fig:pseudoanomaly}(b). At the beginning, $\mathcal{G}$ randomly generates the noise. As $\mathcal{F}$ starts to learn to reconstruct noise-free normal, $\mathcal{G}$ then starts to generate noises around moving objects, where there are many movements so that $\mathcal{F}$ cannot easily remove the noise. 
In this way, $\mathcal{G}$ presents diverse (i.e., from high to low as seen in Figure \ref{fig:trend}(b)) noises to $\mathcal{F}$ across the training iterations and $\mathcal{F}$ finally learns to reconstruct normal regardless of all these kinds of variations given from $\mathcal{G}$. 

We can also observe the enhancement of the reconstruction boundary in Figure \ref{fig:trend}(c) and (d). The rising AUC in Figure \ref{fig:trend}(c) indicates the increased discriminative ability of the AE between normal and anomalous data, reflecting improved reconstruction boundaries. This improvement is further substantiated by the growing disparity between the mean anomaly score of anomalous data, denoted as $\mu^\mathcal{A}_A$, and the mean anomaly score of normal data, denoted as $\mu^\mathcal{A}_N$, as depicted in Figure \ref{fig:trend}(d).


\section{Conclusion}
In this work, we introduced a novel approach for generating pseudo anomalies by incorporating noise into the input without imposing inductive bias. Rather than using a non-learnable noise to generate pseudo anomalies, 
we proposed to utilize an additional autoencoder that learns to generate this noise. 
We provided ablation studies and evaluations using Ped2, Avenue, ShanghaiTech, CIFAR-10, KDDCUP datasets to demonstrate the importance of the noise-based pseudo anomalies and the training mechanism to generate noise. Even without inductive bias, our approach demonstrated superiority and generic applicability to diverse application domains spanning videos, images, and network intrusion while achieving comparable performance to the existing state-of-the-art methods.

\section*{Acknowledgments} 
\noindent This work was supported by Institute of Information \& communications Technology Planning \& Evaluation (IITP) grant funded by the Korea government (MSIT) (RS-2023-00215760, Guide Dog: Development of Navigation AI Technology of a Guidance Robot for the Visually Impaired Person (90\%) and by the Luxembourg National Research Fund (FNR) under the project BRIDGES2021/IS/16353350/FaKeDeTeR (10\%).

\section*{Data availability}
The data that support the findings of this study are publicly available: Ped2 \cite{li2013anomaly} at \url{http://www.svcl.ucsd.edu/projects/anomaly/dataset.htm}; Avenue \cite{lu2013abnormal} at \url{http://www.cse.cuhk.edu.hk/leojia/projects/detectabnormal/dataset.html}; ShanghaiTech \cite{luo2017revisit} at \url{https://svip-lab.github.io/dataset/campus\_dataset.html}; CIFAR-10 \cite{krizhevsky2009learning} at \url{https://www.cs.toronto.edu/~kriz/cifar.html}; and KDDCUP99 \cite{Dua:2019} at \url{http://kdd.ics.uci.edu/databases/kddcup99/kddcup99.html}.

\section*{Conflict of interest}
Marcella Astrid and Muhammad Zaigham Zaheer were recently affiliated and having a close collaboration with the University of Science and Technology, South Korea and the Electronics and Telecommunications Research Institute, South Korea. Marcella Astrid and Djamila Aouada are currently employed at the University of Luxembourg, Luxembourg. Muhammad Zaigham Zaheer is currently employed at the Mohamed Bin Zayed University of Artificial Intelligence, United Arab Emirates. Seung-Ik Lee is currently employed at the University of Science and Technology, South Korea and the Electronics and Telecommunications Research Institute, South Korea.



\bibliography{sn-bibliography}


\begin{thebibliography}{63}
\ifx \bisbn   \undefined \def \bisbn  #1{ISBN #1}\fi
\ifx \binits  \undefined \def \binits#1{#1}\fi
\ifx \bauthor  \undefined \def \bauthor#1{#1}\fi
\ifx \batitle  \undefined \def \batitle#1{#1}\fi
\ifx \bjtitle  \undefined \def \bjtitle#1{#1}\fi
\ifx \bvolume  \undefined \def \bvolume#1{\textbf{#1}}\fi
\ifx \byear  \undefined \def \byear#1{#1}\fi
\ifx \bissue  \undefined \def \bissue#1{#1}\fi
\ifx \bfpage  \undefined \def \bfpage#1{#1}\fi
\ifx \blpage  \undefined \def \blpage #1{#1}\fi
\ifx \burl  \undefined \def \burl#1{\textsf{#1}}\fi
\ifx \doiurl  \undefined \def \doiurl#1{\url{https://doi.org/#1}}\fi
\ifx \betal  \undefined \def \betal{\textit{et al.}}\fi
\ifx \binstitute  \undefined \def \binstitute#1{#1}\fi
\ifx \binstitutionaled  \undefined \def \binstitutionaled#1{#1}\fi
\ifx \bctitle  \undefined \def \bctitle#1{#1}\fi
\ifx \beditor  \undefined \def \beditor#1{#1}\fi
\ifx \bpublisher  \undefined \def \bpublisher#1{#1}\fi
\ifx \bbtitle  \undefined \def \bbtitle#1{#1}\fi
\ifx \bedition  \undefined \def \bedition#1{#1}\fi
\ifx \bseriesno  \undefined \def \bseriesno#1{#1}\fi
\ifx \blocation  \undefined \def \blocation#1{#1}\fi
\ifx \bsertitle  \undefined \def \bsertitle#1{#1}\fi
\ifx \bsnm \undefined \def \bsnm#1{#1}\fi
\ifx \bsuffix \undefined \def \bsuffix#1{#1}\fi
\ifx \bparticle \undefined \def \bparticle#1{#1}\fi
\ifx \barticle \undefined \def \barticle#1{#1}\fi
\bibcommenthead
\ifx \bconfdate \undefined \def \bconfdate #1{#1}\fi
\ifx \botherref \undefined \def \botherref #1{#1}\fi
\ifx \url \undefined \def \url#1{\textsf{#1}}\fi
\ifx \bchapter \undefined \def \bchapter#1{#1}\fi
\ifx \bbook \undefined \def \bbook#1{#1}\fi
\ifx \bcomment \undefined \def \bcomment#1{#1}\fi
\ifx \oauthor \undefined \def \oauthor#1{#1}\fi
\ifx \citeauthoryear \undefined \def \citeauthoryear#1{#1}\fi
\ifx \endbibitem  \undefined \def \endbibitem {}\fi
\ifx \bconflocation  \undefined \def \bconflocation#1{#1}\fi
\ifx \arxivurl  \undefined \def \arxivurl#1{\textsf{#1}}\fi
\csname PreBibitemsHook\endcsname

\bibitem{zaheer2020old}
\begin{bchapter}
\bauthor{\bsnm{Zaheer}, \binits{M.Z.}},
\bauthor{\bsnm{Lee}, \binits{J.-h.}},
\bauthor{\bsnm{Astrid}, \binits{M.}},
\bauthor{\bsnm{Lee}, \binits{S.-I.}}:
\bctitle{Old is gold: Redefining the adversarially learned one-class classifier
  training paradigm}.
In: \bbtitle{Proceedings of the IEEE/CVF Conference on Computer Vision and
  Pattern Recognition},
pp. \bfpage{14183}--\blpage{14193}
(\byear{2020})
\end{bchapter}
\endbibitem

\bibitem{gong2019memorizing}
\begin{bchapter}
\bauthor{\bsnm{Gong}, \binits{D.}},
\bauthor{\bsnm{Liu}, \binits{L.}},
\bauthor{\bsnm{Le}, \binits{V.}},
\bauthor{\bsnm{Saha}, \binits{B.}},
\bauthor{\bsnm{Mansour}, \binits{M.R.}},
\bauthor{\bsnm{Venkatesh}, \binits{S.}},
\bauthor{\bsnm{Hengel}, \binits{A.v.d.}}:
\bctitle{Memorizing normality to detect anomaly: Memory-augmented deep
  autoencoder for unsupervised anomaly detection}.
In: \bbtitle{Proceedings of the IEEE International Conference on Computer
  Vision},
pp. \bfpage{1705}--\blpage{1714}
(\byear{2019})
\end{bchapter}
\endbibitem

\bibitem{li2022anomaly}
\begin{barticle}
\bauthor{\bsnm{Li}, \binits{S.}},
\bauthor{\bsnm{Cheng}, \binits{Y.}},
\bauthor{\bsnm{Tian}, \binits{Y.}},
\bauthor{\bsnm{Liu}, \binits{Y.}}:
\batitle{Anomaly detection based on superpixels in videos}.
\bjtitle{Neural Computing and Applications}
\bvolume{34}(\bissue{15}),
\bfpage{12617}--\blpage{12631}
(\byear{2022})
\end{barticle}
\endbibitem

\bibitem{astrid2021learning}
\begin{bchapter}
\bauthor{\bsnm{Astrid}, \binits{M.}},
\bauthor{\bsnm{Zaheer}, \binits{M.Z.}},
\bauthor{\bsnm{Lee}, \binits{J.-Y.}},
\bauthor{\bsnm{Lee}, \binits{S.-I.}}:
\bctitle{Learning not to reconstruct anomalies}.
In: \bbtitle{British Machine Vision Conference}
(\byear{2021})
\end{bchapter}
\endbibitem

\bibitem{sultani2018real}
\begin{bchapter}
\bauthor{\bsnm{Sultani}, \binits{W.}},
\bauthor{\bsnm{Chen}, \binits{C.}},
\bauthor{\bsnm{Shah}, \binits{M.}}:
\bctitle{Real-world anomaly detection in surveillance videos}.
In: \bbtitle{Proceedings of the IEEE Conference on Computer Vision and Pattern
  Recognition},
pp. \bfpage{6479}--\blpage{6488}
(\byear{2018})
\end{bchapter}
\endbibitem

\bibitem{wahid2021nanod}
\begin{barticle}
\bauthor{\bsnm{Wahid}, \binits{A.}},
\bauthor{\bsnm{Annavarapu}, \binits{C.S.R.}}:
\batitle{Nanod: A natural neighbour-based outlier detection algorithm}.
\bjtitle{Neural Computing and Applications}
\bvolume{33},
\bfpage{2107}--\blpage{2123}
(\byear{2021})
\end{barticle}
\endbibitem

\bibitem{abhaya2022rdpod}
\begin{barticle}
\bauthor{\bsnm{Abhaya}, \binits{A.}},
\bauthor{\bsnm{Patra}, \binits{B.K.}}:
\batitle{Rdpod: an unsupervised approach for outlier detection}.
\bjtitle{Neural Computing and Applications}
\bvolume{34}(\bissue{2}),
\bfpage{1065}--\blpage{1077}
(\byear{2022})
\end{barticle}
\endbibitem

\bibitem{vafaeiflexible}
\begin{botherref}
\oauthor{\bsnm{Vafaei~Sadr}, \binits{A.}},
\oauthor{\bsnm{Bassett}, \binits{B.A.}},
\oauthor{\bsnm{Kunz}, \binits{M.}}:
A flexible framework for anomaly detection via dimensionality reduction.
Neural Computing and Applications,
1--11
\end{botherref}
\endbibitem

\bibitem{zavrak2023flow}
\begin{barticle}
\bauthor{\bsnm{Zavrak}, \binits{S.}},
\bauthor{\bsnm{Iskefiyeli}, \binits{M.}}:
\batitle{Flow-based intrusion detection on software-defined networks: a
  multivariate time series anomaly detection approach}.
\bjtitle{Neural Computing and Applications}
\bvolume{35}(\bissue{16}),
\bfpage{12175}--\blpage{12193}
(\byear{2023})
\end{barticle}
\endbibitem

\bibitem{shukla2021detection}
\begin{barticle}
\bauthor{\bsnm{Shukla}, \binits{A.K.}}:
\batitle{Detection of anomaly intrusion utilizing self-adaptive grasshopper
  optimization algorithm}.
\bjtitle{Neural Computing and Applications}
\bvolume{33}(\bissue{13}),
\bfpage{7541}--\blpage{7561}
(\byear{2021})
\end{barticle}
\endbibitem

\bibitem{saeed2022real}
\begin{barticle}
\bauthor{\bsnm{Saeed}, \binits{M.M.}}:
\batitle{A real-time adaptive network intrusion detection for streaming data: a
  hybrid approach}.
\bjtitle{Neural Computing and Applications}
\bvolume{34}(\bissue{8}),
\bfpage{6227}--\blpage{6240}
(\byear{2022})
\end{barticle}
\endbibitem

\bibitem{luo2017revisit}
\begin{bchapter}
\bauthor{\bsnm{Luo}, \binits{W.}},
\bauthor{\bsnm{Liu}, \binits{W.}},
\bauthor{\bsnm{Gao}, \binits{S.}}:
\bctitle{A revisit of sparse coding based anomaly detection in stacked rnn
  framework}.
In: \bbtitle{Proceedings of the IEEE International Conference on Computer
  Vision},
pp. \bfpage{341}--\blpage{349}
(\byear{2017})
\end{bchapter}
\endbibitem

\bibitem{luo2017remembering}
\begin{bchapter}
\bauthor{\bsnm{Luo}, \binits{W.}},
\bauthor{\bsnm{Liu}, \binits{W.}},
\bauthor{\bsnm{Gao}, \binits{S.}}:
\bctitle{Remembering history with convolutional lstm for anomaly detection}.
In: \bbtitle{IEEE International Conference on Multimedia and Expo},
pp. \bfpage{439}--\blpage{444}
(\byear{2017}).
\bcomment{IEEE}
\end{bchapter}
\endbibitem

\bibitem{park2020learning}
\begin{bchapter}
\bauthor{\bsnm{Park}, \binits{H.}},
\bauthor{\bsnm{Noh}, \binits{J.}},
\bauthor{\bsnm{Ham}, \binits{B.}}:
\bctitle{Learning memory-guided normality for anomaly detection}.
In: \bbtitle{Proceedings of the IEEE/CVF Conference on Computer Vision and
  Pattern Recognition},
pp. \bfpage{14372}--\blpage{14381}
(\byear{2020})
\end{bchapter}
\endbibitem

\bibitem{astrid2021synthetic}
\begin{bchapter}
\bauthor{\bsnm{Astrid}, \binits{M.}},
\bauthor{\bsnm{Zaheer}, \binits{M.Z.}},
\bauthor{\bsnm{Lee}, \binits{S.-I.}}:
\bctitle{Synthetic temporal anomaly guided end-to-end video anomaly detection}.
In: \bbtitle{Proceedings of the IEEE/CVF International Conference on Computer
  Vision Workshops},
pp. \bfpage{207}--\blpage{214}
(\byear{2021})
\end{bchapter}
\endbibitem

\bibitem{liu2023csiamese}
\begin{botherref}
\oauthor{\bsnm{Liu}, \binits{D.}},
\oauthor{\bsnm{Zhong}, \binits{S.}},
\oauthor{\bsnm{Lin}, \binits{L.}},
\oauthor{\bsnm{Zhao}, \binits{M.}},
\oauthor{\bsnm{Fu}, \binits{X.}},
\oauthor{\bsnm{Liu}, \binits{X.}}:
Csiamese: a novel semi-supervised anomaly detection framework for gas turbines
  via reconstruction similarity.
Neural Computing and Applications,
1--25
(2023)
\end{botherref}
\endbibitem

\bibitem{munawar2017limiting}
\begin{bchapter}
\bauthor{\bsnm{Munawar}, \binits{A.}},
\bauthor{\bsnm{Vinayavekhin}, \binits{P.}},
\bauthor{\bsnm{De~Magistris}, \binits{G.}}:
\bctitle{Limiting the reconstruction capability of generative neural network
  using negative learning}.
In: \bbtitle{2017 IEEE 27th International Workshop on Machine Learning for
  Signal Processing (MLSP)},
pp. \bfpage{1}--\blpage{6}
(\byear{2017}).
\bcomment{IEEE}
\end{bchapter}
\endbibitem

\bibitem{voss2007essentials}
\begin{botherref}
\oauthor{\bsnm{Voss}, \binits{P.}}:
Essentials of general intelligence: The direct path to artificial general
  intelligence.
Artificial general intelligence,
131--157
(2007)
\end{botherref}
\endbibitem

\bibitem{li2013anomaly}
\begin{barticle}
\bauthor{\bsnm{Li}, \binits{W.}},
\bauthor{\bsnm{Mahadevan}, \binits{V.}},
\bauthor{\bsnm{Vasconcelos}, \binits{N.}}:
\batitle{Anomaly detection and localization in crowded scenes}.
\bjtitle{IEEE transactions on pattern analysis and machine intelligence}
\bvolume{36}(\bissue{1}),
\bfpage{18}--\blpage{32}
(\byear{2013})
\end{barticle}
\endbibitem

\bibitem{lu2013abnormal}
\begin{bchapter}
\bauthor{\bsnm{Lu}, \binits{C.}},
\bauthor{\bsnm{Shi}, \binits{J.}},
\bauthor{\bsnm{Jia}, \binits{J.}}:
\bctitle{Abnormal event detection at 150 fps in matlab}.
In: \bbtitle{Proceedings of the IEEE International Conference on Computer
  Vision},
pp. \bfpage{2720}--\blpage{2727}
(\byear{2013})
\end{bchapter}
\endbibitem

\bibitem{krizhevsky2009learning}
\begin{botherref}
\oauthor{\bsnm{Krizhevsky}, \binits{A.}},
\oauthor{\bsnm{Hinton}, \binits{G.}}, et al.:
Learning multiple layers of features from tiny images
(2009)
\end{botherref}
\endbibitem

\bibitem{Dua:2019}
\begin{botherref}
\oauthor{\bsnm{Dua}, \binits{D.}},
\oauthor{\bsnm{Graff}, \binits{C.}}:
{UCI} Machine Learning Repository
(2017).
\url{http://archive.ics.uci.edu/ml}
\end{botherref}
\endbibitem

\bibitem{astrid2023pseudobound}
\begin{barticle}
\bauthor{\bsnm{Astrid}, \binits{M.}},
\bauthor{\bsnm{Zaheer}, \binits{M.Z.}},
\bauthor{\bsnm{Lee}, \binits{S.-I.}}:
\batitle{Pseudobound: Limiting the anomaly reconstruction capability of
  one-class classifiers using pseudo anomalies}.
\bjtitle{Neurocomputing}
\bvolume{534},
\bfpage{147}--\blpage{160}
(\byear{2023})
\end{barticle}
\endbibitem

\bibitem{zhong2022cascade}
\begin{barticle}
\bauthor{\bsnm{Zhong}, \binits{Y.}},
\bauthor{\bsnm{Chen}, \binits{X.}},
\bauthor{\bsnm{Jiang}, \binits{J.}},
\bauthor{\bsnm{Ren}, \binits{F.}}:
\batitle{A cascade reconstruction model with generalization ability evaluation
  for anomaly detection in videos}.
\bjtitle{Pattern Recognition}
\bvolume{122},
\bfpage{108336}
(\byear{2022})
\end{barticle}
\endbibitem

\bibitem{goodfellow2014generative}
\begin{botherref}
\oauthor{\bsnm{Goodfellow}, \binits{I.}},
\oauthor{\bsnm{Pouget-Abadie}, \binits{J.}},
\oauthor{\bsnm{Mirza}, \binits{M.}},
\oauthor{\bsnm{Xu}, \binits{B.}},
\oauthor{\bsnm{Warde-Farley}, \binits{D.}},
\oauthor{\bsnm{Ozair}, \binits{S.}},
\oauthor{\bsnm{Courville}, \binits{A.}},
\oauthor{\bsnm{Bengio}, \binits{Y.}}:
Generative adversarial nets.
Advances in neural information processing systems
\textbf{27}
(2014)
\end{botherref}
\endbibitem

\bibitem{zaheer2022stabilizing}
\begin{botherref}
\oauthor{\bsnm{Zaheer}, \binits{M.Z.}},
\oauthor{\bsnm{Lee}, \binits{J.H.}},
\oauthor{\bsnm{Mahmood}, \binits{A.}},
\oauthor{\bsnm{Astrid}, \binits{M.}},
\oauthor{\bsnm{Lee}, \binits{S.-I.}}:
Stabilizing adversarially learned one-class novelty detection using pseudo
  anomalies.
arXiv preprint arXiv:2203.13716
(2022)
\end{botherref}
\endbibitem

\bibitem{pourreza2021g2d}
\begin{bchapter}
\bauthor{\bsnm{Pourreza}, \binits{M.}},
\bauthor{\bsnm{Mohammadi}, \binits{B.}},
\bauthor{\bsnm{Khaki}, \binits{M.}},
\bauthor{\bsnm{Bouindour}, \binits{S.}},
\bauthor{\bsnm{Snoussi}, \binits{H.}},
\bauthor{\bsnm{Sabokrou}, \binits{M.}}:
\bctitle{G2d: Generate to detect anomaly}.
In: \bbtitle{Proceedings of the IEEE/CVF Winter Conference on Applications of
  Computer Vision},
pp. \bfpage{2003}--\blpage{2012}
(\byear{2021})
\end{bchapter}
\endbibitem

\bibitem{liu2018future}
\begin{bchapter}
\bauthor{\bsnm{Liu}, \binits{W.}},
\bauthor{\bsnm{Luo}, \binits{W.}},
\bauthor{\bsnm{Lian}, \binits{D.}},
\bauthor{\bsnm{Gao}, \binits{S.}}:
\bctitle{Future frame prediction for anomaly detection--a new baseline}.
In: \bbtitle{Proceedings of the IEEE Conference on Computer Vision and Pattern
  Recognition},
pp. \bfpage{6536}--\blpage{6545}
(\byear{2018})
\end{bchapter}
\endbibitem

\bibitem{ionescu2019object}
\begin{bchapter}
\bauthor{\bsnm{Ionescu}, \binits{R.T.}},
\bauthor{\bsnm{Khan}, \binits{F.S.}},
\bauthor{\bsnm{Georgescu}, \binits{M.-I.}},
\bauthor{\bsnm{Shao}, \binits{L.}}:
\bctitle{Object-centric auto-encoders and dummy anomalies for abnormal event
  detection in video}.
In: \bbtitle{Proceedings of the IEEE/CVF Conference on Computer Vision and
  Pattern Recognition},
pp. \bfpage{7842}--\blpage{7851}
(\byear{2019})
\end{bchapter}
\endbibitem

\bibitem{georgescu2021background}
\begin{botherref}
\oauthor{\bsnm{Georgescu}, \binits{M.I.}},
\oauthor{\bsnm{Ionescu}, \binits{R.}},
\oauthor{\bsnm{Khan}, \binits{F.S.}},
\oauthor{\bsnm{Popescu}, \binits{M.}},
\oauthor{\bsnm{Shah}, \binits{M.}}:
A background-agnostic framework with adversarial training for abnormal event
  detection in video.
IEEE Transactions on Pattern Analysis \& Machine Intelligence
(01),
1--1
(2021)
\end{botherref}
\endbibitem

\bibitem{ji2020tam}
\begin{bchapter}
\bauthor{\bsnm{Ji}, \binits{X.}},
\bauthor{\bsnm{Li}, \binits{B.}},
\bauthor{\bsnm{Zhu}, \binits{Y.}}:
\bctitle{Tam-net: Temporal enhanced appearance-to-motion generative network for
  video anomaly detection}.
In: \bbtitle{2020 International Joint Conference on Neural Networks (IJCNN)},
pp. \bfpage{1}--\blpage{8}
(\byear{2020}).
\bcomment{IEEE}
\end{bchapter}
\endbibitem

\bibitem{lee2019bman}
\begin{barticle}
\bauthor{\bsnm{Lee}, \binits{S.}},
\bauthor{\bsnm{Kim}, \binits{H.G.}},
\bauthor{\bsnm{Ro}, \binits{Y.M.}}:
\batitle{Bman: Bidirectional multi-scale aggregation networks for abnormal
  event detection}.
\bjtitle{IEEE Transactions on Image Processing}
\bvolume{29},
\bfpage{2395}--\blpage{2408}
(\byear{2019})
\end{barticle}
\endbibitem

\bibitem{yamanaka2019autoencoding}
\begin{bchapter}
\bauthor{\bsnm{Yamanaka}, \binits{Y.}},
\bauthor{\bsnm{Iwata}, \binits{T.}},
\bauthor{\bsnm{Takahashi}, \binits{H.}},
\bauthor{\bsnm{Yamada}, \binits{M.}},
\bauthor{\bsnm{Kanai}, \binits{S.}}:
\bctitle{Autoencoding binary classifiers for supervised anomaly detection}.
In: \bbtitle{Pacific Rim International Conference on Artificial Intelligence},
pp. \bfpage{647}--\blpage{659}
(\byear{2019}).
\bcomment{Springer}
\end{bchapter}
\endbibitem

\bibitem{zaheer2020claws}
\begin{bchapter}
\bauthor{\bsnm{Zaheer}, \binits{M.Z.}},
\bauthor{\bsnm{Mahmood}, \binits{A.}},
\bauthor{\bsnm{Astrid}, \binits{M.}},
\bauthor{\bsnm{Lee}, \binits{S.-I.}}:
\bctitle{Claws: Clustering assisted weakly supervised learning with normalcy
  suppression for anomalous event detection}.
In: \bbtitle{Proceedings of the European Conference on Computer Vision (ECCV)}
(\byear{2020})
\end{bchapter}
\endbibitem

\bibitem{zaheer2023clustering}
\begin{botherref}
\oauthor{\bsnm{Zaheer}, \binits{M.Z.}},
\oauthor{\bsnm{Mahmood}, \binits{A.}},
\oauthor{\bsnm{Astrid}, \binits{M.}},
\oauthor{\bsnm{Lee}, \binits{S.-I.}}:
Clustering aided weakly supervised training to detect anomalous events in
  surveillance videos.
IEEE Transactions on Neural Networks and Learning Systems
(2023)
\end{botherref}
\endbibitem

\bibitem{karim2024real}
\begin{bchapter}
\bauthor{\bsnm{Karim}, \binits{H.}},
\bauthor{\bsnm{Doshi}, \binits{K.}},
\bauthor{\bsnm{Yilmaz}, \binits{Y.}}:
\bctitle{Real-time weakly supervised video anomaly detection}.
In: \bbtitle{Proceedings of the IEEE/CVF Winter Conference on Applications of
  Computer Vision},
pp. \bfpage{6848}--\blpage{6856}
(\byear{2024})
\end{bchapter}
\endbibitem

\bibitem{majhi2024oe}
\begin{bchapter}
\bauthor{\bsnm{Majhi}, \binits{S.}},
\bauthor{\bsnm{Dai}, \binits{R.}},
\bauthor{\bsnm{Kong}, \binits{Q.}},
\bauthor{\bsnm{Garattoni}, \binits{L.}},
\bauthor{\bsnm{Francesca}, \binits{G.}},
\bauthor{\bsnm{Br{\'e}mond}, \binits{F.}}:
\bctitle{Oe-ctst: Outlier-embedded cross temporal scale transformer for
  weakly-supervised video anomaly detection}.
In: \bbtitle{Proceedings of the IEEE/CVF Winter Conference on Applications of
  Computer Vision},
pp. \bfpage{8574}--\blpage{8583}
(\byear{2024})
\end{bchapter}
\endbibitem

\bibitem{zaheer2022generative}
\begin{bchapter}
\bauthor{\bsnm{Zaheer}, \binits{M.Z.}},
\bauthor{\bsnm{Mahmood}, \binits{A.}},
\bauthor{\bsnm{Khan}, \binits{M.H.}},
\bauthor{\bsnm{Segu}, \binits{M.}},
\bauthor{\bsnm{Yu}, \binits{F.}},
\bauthor{\bsnm{Lee}, \binits{S.-I.}}:
\bctitle{Generative cooperative learning for unsupervised video anomaly
  detection}.
In: \bbtitle{Proceedings of the IEEE/CVF Conference on Computer Vision and
  Pattern Recognition},
pp. \bfpage{14744}--\blpage{14754}
(\byear{2022})
\end{bchapter}
\endbibitem

\bibitem{ionescu2019detecting}
\begin{bchapter}
\bauthor{\bsnm{Ionescu}, \binits{R.T.}},
\bauthor{\bsnm{Smeureanu}, \binits{S.}},
\bauthor{\bsnm{Popescu}, \binits{M.}},
\bauthor{\bsnm{Alexe}, \binits{B.}}:
\bctitle{Detecting abnormal events in video using narrowed normality clusters}.
In: \bbtitle{2019 IEEE Winter Conference on Applications of Computer Vision
  (WACV)},
pp. \bfpage{1951}--\blpage{1960}
(\byear{2019}).
\bcomment{IEEE}
\end{bchapter}
\endbibitem

\bibitem{salehi2021arae}
\begin{barticle}
\bauthor{\bsnm{Salehi}, \binits{M.}},
\bauthor{\bsnm{Arya}, \binits{A.}},
\bauthor{\bsnm{Pajoum}, \binits{B.}},
\bauthor{\bsnm{Otoofi}, \binits{M.}},
\bauthor{\bsnm{Shaeiri}, \binits{A.}},
\bauthor{\bsnm{Rohban}, \binits{M.H.}},
\bauthor{\bsnm{Rabiee}, \binits{H.R.}}:
\batitle{Arae: Adversarially robust training of autoencoders improves novelty
  detection}.
\bjtitle{Neural Networks}
\bvolume{144},
\bfpage{726}--\blpage{736}
(\byear{2021})
\end{barticle}
\endbibitem

\bibitem{jewell2022one}
\begin{bchapter}
\bauthor{\bsnm{Jewell}, \binits{J.T.}},
\bauthor{\bsnm{Khazaie}, \binits{V.R.}},
\bauthor{\bsnm{Mohsenzadeh}, \binits{Y.}}:
\bctitle{One-class learned encoder-decoder network with adversarial context
  masking for novelty detection}.
In: \bbtitle{Proceedings of the IEEE/CVF Winter Conference on Applications of
  Computer Vision},
pp. \bfpage{3591}--\blpage{3601}
(\byear{2022})
\end{bchapter}
\endbibitem

\bibitem{vincent2008extracting}
\begin{bchapter}
\bauthor{\bsnm{Vincent}, \binits{P.}},
\bauthor{\bsnm{Larochelle}, \binits{H.}},
\bauthor{\bsnm{Bengio}, \binits{Y.}},
\bauthor{\bsnm{Manzagol}, \binits{P.-A.}}:
\bctitle{Extracting and composing robust features with denoising autoencoders}.
In: \bbtitle{Proceedings of the 25th International Conference on Machine
  Learning},
pp. \bfpage{1096}--\blpage{1103}
(\byear{2008})
\end{bchapter}
\endbibitem

\bibitem{bengio2011deep}
\begin{bchapter}
\bauthor{\bsnm{Bengio}, \binits{Y.}},
\bauthor{\bsnm{Bastien}, \binits{F.}},
\bauthor{\bsnm{Bergeron}, \binits{A.}},
\bauthor{\bsnm{Boulanger--Lewandowski}, \binits{N.}},
\bauthor{\bsnm{Breuel}, \binits{T.}},
\bauthor{\bsnm{Chherawala}, \binits{Y.}},
\bauthor{\bsnm{Cisse}, \binits{M.}},
\bauthor{\bsnm{C{\^o}t{\'e}}, \binits{M.}},
\bauthor{\bsnm{Erhan}, \binits{D.}},
\bauthor{\bsnm{Eustache}, \binits{J.}}, \betal:
\bctitle{Deep learners benefit more from out-of-distribution examples}.
In: \bbtitle{Proceedings of the 14th International Conference on Artificial
  Intelligence and Statistics},
pp. \bfpage{164}--\blpage{172}
(\byear{2011}).
\bcomment{JMLR Workshop and Conference Proceedings}
\end{bchapter}
\endbibitem

\bibitem{krizhevsky2012imagenet}
\begin{barticle}
\bauthor{\bsnm{Krizhevsky}, \binits{A.}},
\bauthor{\bsnm{Sutskever}, \binits{I.}},
\bauthor{\bsnm{Hinton}, \binits{G.E.}}:
\batitle{Imagenet classification with deep convolutional neural networks}.
\bjtitle{Advances in neural information processing systems}
\bvolume{25},
\bfpage{1097}--\blpage{1105}
(\byear{2012})
\end{barticle}
\endbibitem

\bibitem{zhang2020adversarial}
\begin{bchapter}
\bauthor{\bsnm{Zhang}, \binits{X.}},
\bauthor{\bsnm{Wang}, \binits{Q.}},
\bauthor{\bsnm{Zhang}, \binits{J.}},
\bauthor{\bsnm{Zhong}, \binits{Z.}}:
\bctitle{Adversarial autoaugment}.
In: \bbtitle{International Conference on Learning Representations}
(\byear{2020}).
\burl{https://openreview.net/forum?id=ByxdUySKvS}
\end{bchapter}
\endbibitem

\bibitem{tang2020onlineaugment}
\begin{bchapter}
\bauthor{\bsnm{Tang}, \binits{Z.}},
\bauthor{\bsnm{Gao}, \binits{Y.}},
\bauthor{\bsnm{Karlinsky}, \binits{L.}},
\bauthor{\bsnm{Sattigeri}, \binits{P.}},
\bauthor{\bsnm{Feris}, \binits{R.}},
\bauthor{\bsnm{Metaxas}, \binits{D.}}:
\bctitle{Onlineaugment: Online data augmentation with less domain knowledge}.
In: \beditor{\bsnm{Vedaldi}, \binits{A.}},
\beditor{\bsnm{Bischof}, \binits{H.}},
\beditor{\bsnm{Brox}, \binits{T.}},
\beditor{\bsnm{Frahm}, \binits{J.-M.}} (eds.)
\bbtitle{Computer Vision -- ECCV 2020},
pp. \bfpage{313}--\blpage{329}.
\bpublisher{Springer},
\blocation{Cham}
(\byear{2020})
\end{bchapter}
\endbibitem

\bibitem{dai2017good}
\begin{botherref}
\oauthor{\bsnm{Dai}, \binits{Z.}},
\oauthor{\bsnm{Yang}, \binits{Z.}},
\oauthor{\bsnm{Yang}, \binits{F.}},
\oauthor{\bsnm{Cohen}, \binits{W.W.}},
\oauthor{\bsnm{Salakhutdinov}, \binits{R.R.}}:
Good semi-supervised learning that requires a bad gan.
Advances in neural information processing systems
\textbf{30}
(2017)
\end{botherref}
\endbibitem

\bibitem{dong2019margingan}
\begin{botherref}
\oauthor{\bsnm{Dong}, \binits{J.}},
\oauthor{\bsnm{Lin}, \binits{T.}}:
Margingan: Adversarial training in semi-supervised learning.
Advances in neural information processing systems
\textbf{32}
(2019)
\end{botherref}
\endbibitem

\bibitem{salimans2017pixelcnn++}
\begin{botherref}
\oauthor{\bsnm{Salimans}, \binits{T.}},
\oauthor{\bsnm{Karpathy}, \binits{A.}},
\oauthor{\bsnm{Chen}, \binits{X.}},
\oauthor{\bsnm{Kingma}, \binits{D.P.}}:
Pixelcnn++: Improving the pixelcnn with discretized logistic mixture likelihood
  and other modifications.
arXiv preprint arXiv:1701.05517
(2017)
\end{botherref}
\endbibitem

\bibitem{du2022towards}
\begin{bchapter}
\bauthor{\bsnm{Du}, \binits{X.}},
\bauthor{\bsnm{Wang}, \binits{Z.}},
\bauthor{\bsnm{Cai}, \binits{M.}},
\bauthor{\bsnm{Li}, \binits{S.}}:
\bctitle{Towards unknown-aware learning with virtual outlier synthesis}.
In: \bbtitle{International Conference on Learning Representations}
(\byear{2022}).
\burl{https://openreview.net/forum?id=TW7d65uYu5M}
\end{bchapter}
\endbibitem

\bibitem{fxray}
\begin{botherref}
\oauthor{\bsnm{Li}, \binits{L.}},
\oauthor{\bsnm{Bao}, \binits{J.}},
\oauthor{\bsnm{Zhang}, \binits{T.}},
\oauthor{\bsnm{Yang}, \binits{H.}},
\oauthor{\bsnm{Chen}, \binits{D.}},
\oauthor{\bsnm{Wen}, \binits{F.}},
\oauthor{\bsnm{Guo}, \binits{B.}}:
Face x-ray for more general face forgery detection.
CoRR
\textbf{abs/1912.13458}
(2019)
{\href{https://arxiv.org/abs/1912.13458}{{1912.13458}}}
\end{botherref}
\endbibitem

\bibitem{sbis}
\begin{bchapter}
\bauthor{\bsnm{Shiohara}, \binits{K.}},
\bauthor{\bsnm{Yamasaki}, \binits{T.}}:
\bctitle{Detecting deepfakes with self-blended images}.
In: \bbtitle{Proceedings of the IEEE/CVF Conference on Computer Vision and
  Pattern Recognition},
pp. \bfpage{18720}--\blpage{18729}
(\byear{2022})
\end{bchapter}
\endbibitem

\bibitem{hasan2016learning}
\begin{bchapter}
\bauthor{\bsnm{Hasan}, \binits{M.}},
\bauthor{\bsnm{Choi}, \binits{J.}},
\bauthor{\bsnm{Neumann}, \binits{J.}},
\bauthor{\bsnm{Roy-Chowdhury}, \binits{A.K.}},
\bauthor{\bsnm{Davis}, \binits{L.S.}}:
\bctitle{Learning temporal regularity in video sequences}.
In: \bbtitle{Proceedings of the IEEE Conference on Computer Vision and Pattern
  Recognition},
pp. \bfpage{733}--\blpage{742}
(\byear{2016})
\end{bchapter}
\endbibitem

\bibitem{zhao2017spatio}
\begin{bchapter}
\bauthor{\bsnm{Zhao}, \binits{Y.}},
\bauthor{\bsnm{Deng}, \binits{B.}},
\bauthor{\bsnm{Shen}, \binits{C.}},
\bauthor{\bsnm{Liu}, \binits{Y.}},
\bauthor{\bsnm{Lu}, \binits{H.}},
\bauthor{\bsnm{Hua}, \binits{X.-S.}}:
\bctitle{Spatio-temporal autoencoder for video anomaly detection}.
In: \bbtitle{Proceedings of the 25th ACM International Conference on
  Multimedia},
pp. \bfpage{1933}--\blpage{1941}
(\byear{2017})
\end{bchapter}
\endbibitem

\bibitem{abati2019latent}
\begin{bchapter}
\bauthor{\bsnm{Abati}, \binits{D.}},
\bauthor{\bsnm{Porrello}, \binits{A.}},
\bauthor{\bsnm{Calderara}, \binits{S.}},
\bauthor{\bsnm{Cucchiara}, \binits{R.}}:
\bctitle{Latent space autoregression for novelty detection}.
In: \bbtitle{Proceedings of the IEEE Conference on Computer Vision and Pattern
  Recognition},
pp. \bfpage{481}--\blpage{490}
(\byear{2019})
\end{bchapter}
\endbibitem

\bibitem{zong2018deep}
\begin{bchapter}
\bauthor{\bsnm{Zong}, \binits{B.}},
\bauthor{\bsnm{Song}, \binits{Q.}},
\bauthor{\bsnm{Min}, \binits{M.R.}},
\bauthor{\bsnm{Cheng}, \binits{W.}},
\bauthor{\bsnm{Lumezanu}, \binits{C.}},
\bauthor{\bsnm{Cho}, \binits{D.}},
\bauthor{\bsnm{Chen}, \binits{H.}}:
\bctitle{Deep autoencoding gaussian mixture model for unsupervised anomaly
  detection}.
In: \bbtitle{International Conference on Learning Representations}
(\byear{2018})
\end{bchapter}
\endbibitem

\bibitem{kingma2014adam}
\begin{botherref}
\oauthor{\bsnm{Kingma}, \binits{D.P.}},
\oauthor{\bsnm{Ba}, \binits{J.}}:
Adam: A method for stochastic optimization.
arXiv preprint arXiv:1412.6980
(2014)
\end{botherref}
\endbibitem

\bibitem{georgescu2021anomaly}
\begin{bchapter}
\bauthor{\bsnm{Georgescu}, \binits{M.-I.}},
\bauthor{\bsnm{Barbalau}, \binits{A.}},
\bauthor{\bsnm{Ionescu}, \binits{R.T.}},
\bauthor{\bsnm{Khan}, \binits{F.S.}},
\bauthor{\bsnm{Popescu}, \binits{M.}},
\bauthor{\bsnm{Shah}, \binits{M.}}:
\bctitle{Anomaly detection in video via self-supervised and multi-task
  learning}.
In: \bbtitle{Proceedings of the IEEE/CVF Conference on Computer Vision and
  Pattern Recognition},
pp. \bfpage{12742}--\blpage{12752}
(\byear{2021})
\end{bchapter}
\endbibitem

\bibitem{astrid2022limiting}
\begin{bchapter}
\bauthor{\bsnm{Astrid}, \binits{M.}},
\bauthor{\bsnm{Zaheer}, \binits{M.Z.}},
\bauthor{\bsnm{Lee}, \binits{S.-I.}}:
\bctitle{Limiting reconstruction capability of autoencoders using moving
  backward pseudo anomalies}.
In: \bbtitle{2022 19th International Conference on Ubiquitous Robots (UR)},
pp. \bfpage{248}--\blpage{251}
(\byear{2022}).
\bcomment{IEEE}
\end{bchapter}
\endbibitem

\bibitem{lu2020few}
\begin{bchapter}
\bauthor{\bsnm{Lu}, \binits{Y.}},
\bauthor{\bsnm{Yu}, \binits{F.}},
\bauthor{\bsnm{Reddy}, \binits{M.K.K.}},
\bauthor{\bsnm{Wang}, \binits{Y.}}:
\bctitle{Few-shot scene-adaptive anomaly detection}.
In: \bbtitle{European Conference on Computer Vision},
pp. \bfpage{125}--\blpage{141}
(\byear{2020}).
\bcomment{Springer}
\end{bchapter}
\endbibitem

\bibitem{wang2023memory}
\begin{botherref}
\oauthor{\bsnm{Wang}, \binits{L.}},
\oauthor{\bsnm{Tian}, \binits{J.}},
\oauthor{\bsnm{Zhou}, \binits{S.}},
\oauthor{\bsnm{Shi}, \binits{H.}},
\oauthor{\bsnm{Hua}, \binits{G.}}:
Memory-augmented appearance-motion network for video anomaly detection.
Pattern Recognition,
109335
(2023)
\end{botherref}
\endbibitem

\bibitem{vu2019robust}
\begin{bchapter}
\bauthor{\bsnm{Vu}, \binits{H.}},
\bauthor{\bsnm{Nguyen}, \binits{T.D.}},
\bauthor{\bsnm{Le}, \binits{T.}},
\bauthor{\bsnm{Luo}, \binits{W.}},
\bauthor{\bsnm{Phung}, \binits{D.}}:
\bctitle{Robust anomaly detection in videos using multilevel representations}.
In: \bbtitle{Proceedings of the AAAI Conference on Artificial Intelligence},
vol. \bseriesno{33},
pp. \bfpage{5216}--\blpage{5223}
(\byear{2019})
\end{bchapter}
\endbibitem

\bibitem{sun2023hierarchical}
\begin{bchapter}
\bauthor{\bsnm{Sun}, \binits{S.}},
\bauthor{\bsnm{Gong}, \binits{X.}}:
\bctitle{Hierarchical semantic contrast for scene-aware video anomaly
  detection}.
In: \bbtitle{Proceedings of the IEEE/CVF Conference on Computer Vision and
  Pattern Recognition},
pp. \bfpage{22846}--\blpage{22856}
(\byear{2023})
\end{bchapter}
\endbibitem

\end{thebibliography}


\end{document}